\documentclass{article}

\PassOptionsToPackage{numbers, compress}{natbib}

\usepackage[main, final]{neurips_2026}

\usepackage[utf8]{inputenc}
\usepackage[T1]{fontenc}
\usepackage{hyperref}
\usepackage{url}
\usepackage{amsfonts}
\usepackage{booktabs} 
\usepackage{multirow}
\usepackage{graphicx}
\usepackage{amsmath}
\usepackage{nicefrac}
\usepackage{microtype}
\usepackage[table]{xcolor}
\usepackage{xspace}
\usepackage{subcaption}
\usepackage{pifont}
\usepackage{float}
\newcommand{\ie}{{\emph{i.e.}},\xspace}
\newcommand{\eg}{{\emph{e.g.}},\xspace}
\newcommand{\cmark}{\ding{51}}
\newcommand{\xmark}{\ding{55}}
\definecolor{custompink}{RGB}{255, 20, 147} 
\hypersetup{
    colorlinks=true, 
    linkcolor=custompink, 
    filecolor=custompink,
    urlcolor=custompink
}

\title{Unaligned Image Guided Denoising via \\ Cross-modal Conditional Flow Matching}

\author{
	Runmin Zhang$^{1}$\thanks{Equal contribution.}\quad
    Linshan Li$^{1}$\footnotemark[1]\quad
    Si-Yuan Cao$^{1}$\quad
    Zhu Yu$^{1}$\\
    \textbf{Lingyu Zhu}$^{2}$\quad
    \textbf{Huaqi Zhang}$^{3}$\quad
    \textbf{Hui-Liang Shen}$^{1}$\thanks{Corresponding author.}\\
	{\small{$^{1}$Zhejiang University}}\quad
    {\small{$^{2}$City University of Hong Kong}}\quad
    {\small{$^{3}$vivo Mobile Communication Co., Ltd.}}\\
	{\tt\small \{runmin\_zhang, linshan\_li, cao\_siyuan, yu\_zhu\}@zju.edu.cn} \\
	{\tt\small lingyzhu-c@my.cityu.edu.hk, zhanghuaqi@vivo.com, shenhl@zju.edu.cn}
}

\newsavebox{\ugdfmtablebox}
\newcommand{\fitTableToWidth}[1]{
    \sbox{\ugdfmtablebox}{#1}
    \ifdim\wd\ugdfmtablebox>\linewidth
        \resizebox{\linewidth}{!}{\usebox{\ugdfmtablebox}}
    \else
        \usebox{\ugdfmtablebox}
    \fi
}

\begin{document}

\maketitle

\begin{abstract}
This work presents UGD-FM, a novel unaligned image guided denoising framework based on cross-modal conditional flow matching. Unlike previous approaches that assume spatially aligned inputs or only handle small-baseline rectified pairs, we address a more challenging setting where severe target noise and large-range two-dimensional cross-modal misalignment coexist. Specifically, we formulate guided denoising as a progressive flow matching process, allowing image restoration and aligned guidance aggregation to mutually reinforce each other. To support efficient few-step inference, we introduce a stage-focused and inference-consistent training strategy that focuses on task-critical denoising stages, and mitigates the training-inference mismatch. We further propose adaptive sparse guidance propagation for efficient and reliable large-range matching, regularized by perceptual and geometric auxiliary losses. Experiments demonstrate that UGD-FM achieves state-of-the-art denoising performance with only one or two inference steps, and is compatible with modern single-image restoration backbones. Code is available at \href{https://github.com/RM-Zhang/UGD-FM}{https://github.com/RM-Zhang/UGD-FM}.
\end{abstract}

\section{Introduction}
Image denoising is a fundamental problem in low-level vision. Although modern restoration networks~\cite{MPRNet_CVPR21, Uformer_CVPR22, Restormer_CVPR22, NAFNet_ECCV22, Xformer_ICLR24} have substantially improved denoising performance, faithfully recovering fine structures from heavily corrupted observations without over-smoothing remains challenging. To alleviate this limitation, recent works~\cite{CUNet_TPAMI20, DVN_AAAI22, FGDNet_TMM22, MNNet_IF22, RFFNet_MM24, SANet_CVPR23, SGDFormer_IF25, FCENet_CVPR25} have leveraged auxiliary guidance images, such as near-infrared (NIR) images, which provide cleaner structural cues for denoising.

However, most guided denoising approaches~\cite{CUNet_TPAMI20, FGDNet_TMM22, MNNet_IF22, RFFNet_MM24, FCENet_CVPR25} assume that the noisy RGB image and the guidance image are spatially aligned. This assumption is difficult to satisfy in practical mobile imaging systems, where cross-modal images are typically captured by different cameras~\cite{CSStereo_CVPR18, MS2_CVPR23}. Recent approaches~\cite{SANet_CVPR23, SGDFormer_IF25} have extended guided denoising to cross-modal stereo pairs, but remain limited to horizontally rectified pairs with small baselines. In contrast, we study unaligned guided denoising under \textbf{more severe spatial misalignment} than the small-baseline rectified stereo setting considered in prior work. This expands the correspondence search from a one-dimensional horizontal search to a two-dimensional search space~\cite{CrossModalRAFT_AAAI22, MINIMA_CVPR25, DCFlow_ICLR26}, making reliable cross-modal matching and guidance extraction significantly more difficult. This challenge is further amplified by severe noise in the target RGB image and inherent cross-modal appearance differences.

\begin{figure}[t]
    \centering
    \includegraphics[width=0.95\linewidth]{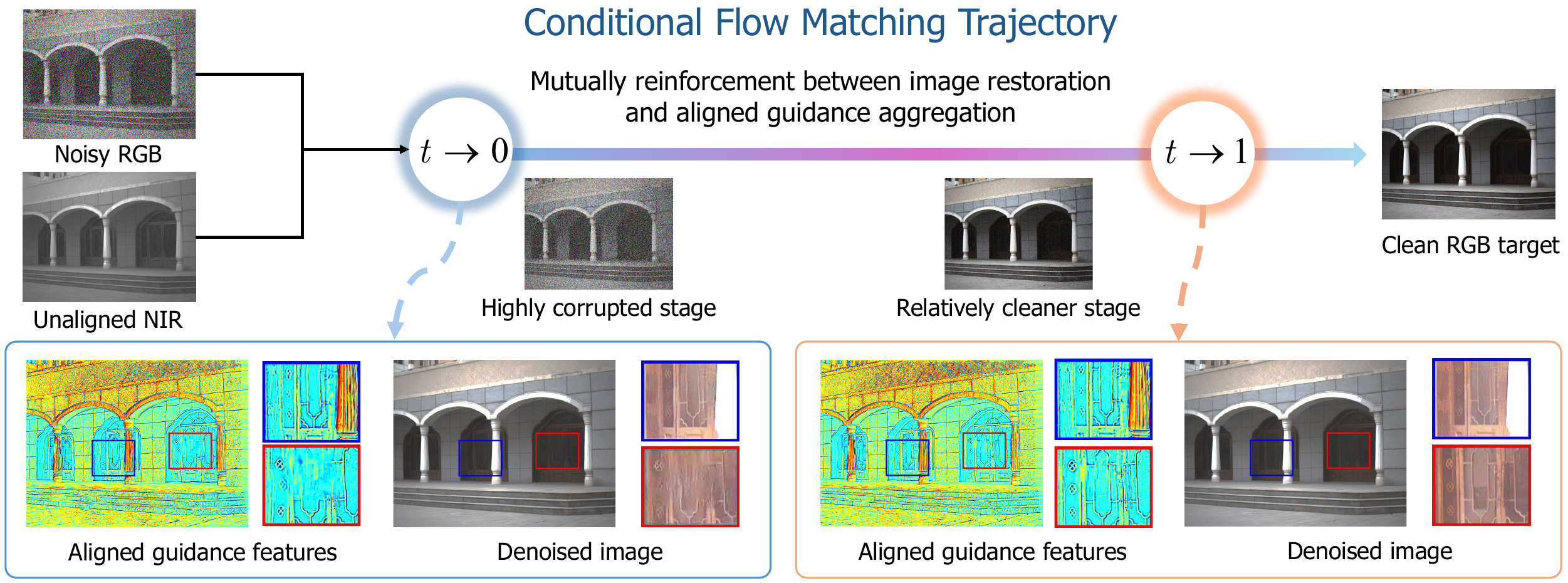}
    \caption{Illustration of the cross-modal conditional flow matching trajectory of UGD-FM. Given a noisy RGB image and an unaligned NIR guidance image, UGD-FM progressively restores the RGB state toward the clean target, forming a mutual reinforcing process between image restoration and guidance aggregation. For clearer visualization, the brightness of the zoom-in patches is enhanced.}
    \label{fig:teaser}
\end{figure}

To address these challenges, we propose UGD-FM, a novel framework for \textbf{U}naligned image \textbf{G}uided \textbf{D}enoising via cross-modal conditional \textbf{F}low \textbf{M}atching. Specifically, we formulate guided denoising as a conditional flow matching process that progressively transports the noisy RGB image distribution toward the clean RGB image distribution, conditioned on unaligned cross-modal guidance. As the transport process progresses, the target image becomes cleaner, making cross-modal correspondence easier to establish and enabling more effective use of the guidance information. As illustrated in Fig.~\ref{fig:teaser}, this creates a mutually reinforcing process between guidance extraction and image restoration, leading to more robust denoising under severe spatial misalignment and noise.

To enable efficient and effective few-step guided denoising, we further design a \textbf{stage-focused and inference-consistent training strategy}. Our key observation is that the denoising trajectory contains two critical stages, \ie a \textbf{highly corrupted stage} for coarse RGB structure recovery, and a \textbf{relatively cleaner stage} for reliable cross-modal matching and fine-detail restoration. Based on this observation, we place the inference timesteps around these stages, and align training with this schedule through stage-focused time sampling and inference-consistent state construction. With this design, UGD-FM delivers strong performance with only a single inference step, and achieves state-of-the-art results with two-step inference.

As for cross-modal matching and guidance extraction, a straightforward solution is to perform dense correspondence search over a large displacement range~\cite{SANet_CVPR23, SGDFormer_IF25}. However, such a dense matching strategy incurs prohibitive computational cost. To mitigate this issue, we introduce an \textbf{adaptive sparse guidance propagation} mechanism. Specifically, our method progressively focuses on a small set of high-confidence correspondence candidates, and propagates reliable guidance cues in a coarse-to-fine manner. To stabilize sparse guidance aggregation, we further introduce two auxiliary losses that provide perceptual and geometric regularization. 

Importantly, we highlight that both our adaptive sparse guidance propagation mechanism and conditional flow matching formulation are compatible with a broad range of modern single-image restoration backbones, such as NAFNet~\cite{NAFNet_ECCV22}, Uformer~\cite{Uformer_CVPR22}, and Restormer~\cite{Restormer_CVPR22}. Our contributions are summarized as follows:
\begin{itemize}
    \item We propose UGD-FM, a general cross-modal conditional flow matching framework for unaligned image guided denoising. UGD-FM establishes a mutually reinforcing process between image restoration and guidance aggregation, enabling robust restoration under severe noise and large-range spatial misalignment.

    \item We design a stage-focused and inference-consistent training strategy for few-step inference. It focuses inference and training on two critical denoising stages, and mitigates the training-inference mismatch, enabling effective one-step and two-step guided denoising.

    \item We introduce adaptive sparse guidance propagation for efficient large-range cross-modal matching and aligned guidance aggregation, regularized by perceptual and geometric auxiliary losses to stabilize sparse correspondence learning.

    \item Comprehensive experiments demonstrate that UGD-FM achieves state-of-the-art denoising performance, and is compatible with modern single-image restoration backbones.
\end{itemize}

\section{Related Work}
\textbf{Single Image Restoration.} Single image restoration aims to recover clean images from degraded observations. Recently, deep convolutional neural networks have become a dominant paradigm by learning mappings from degraded inputs to clean targets. Representative designs include residual learning~\cite{DnCNN_TIP17}, multi-stage refinement~\cite{MPRNet_CVPR21}, and inter-/intra-block architectures~\cite{NAFNet_ECCV22}. Vision transformers and their variants~\cite{IPT_CVPR21,SwinIR_ICCV2021,Uformer_CVPR22,Restormer_CVPR22,Xformer_ICLR24} further improve restoration performance by modeling long-range context. Despite these advances, single image restoration remains fundamentally ill-posed under severe degradation, often leading to over-smoothed results and missing structural details.

\textbf{Guided Image Restoration.} Guided image restoration leverages guidance images from auxiliary modalities to provide complementary structural information, with NIR-guided RGB denoising being a representative application. Early methods~\cite{GF_TPAMI12,DJF_ECCV2016,SVLRM_CVPR19} employ joint filtering to model structural relationships between target and guidance images. Further approaches improve denoising performance with convolutional sparse coding~\cite{CUNet_TPAMI20}, ADMM-based optimization~\cite{MNNet_IF22}, frequency decomposition~\cite{FGDNet_TMM22,FRDN_ICIP22,FCENet_CVPR25}, and multi-modal inconsistency priors~\cite{DVN_AAAI22,RFFNet_MM24,NAID_TMM24}. Nevertheless, most of these approaches assume that the target and guidance images are strictly aligned. This requirement usually depends on specialized imaging systems or careful calibration~\cite{FANVID_ECCV24,SANet_CVPR23,RNS_CVPR25}, making it difficult to satisfy in practical mobile imaging systems. Recent approaches~\cite{MJF_ICCV17,RedFlash_CVPR21,SSC-HSR_IJCV2025,3MTI_CVPR2026,SANet_CVPR23,SGDFormer_IF25} start to relax the strict alignment assumption. However, they either depend on external optical flow networks for pre-alignment, or handle limited misalignment. Other approaches~\cite{PromptFusion_JAS24, DCEvo_CVPR25, UniFusion_CVPR26, DRFusion_Arxiv2605, SFRF_Arxiv2605} handle infrared and visible image fusion, in which SFRF~\cite{SFRF_Arxiv2605} introduces multi-scale iterative registration. In contrast, we address a more challenging setting, where severe target noise and large-range cross-modal displacement must be addressed jointly.

\textbf{Generative Models for Image Restoration.} Generative priors, particularly diffusion~\cite{3MTI_CVPR2026, DiffPIR_CVPR2023, FGPS_ICCV2025, DiFiisr_CVPR25} and flow matching~\cite{RAW-Flow_AAAI2026,RFfusion_NIPS2025,ResFlow_CVPR25} models, have demonstrated strong capabilities in image restoration by learning to transport a simple base distribution to the data manifold. Recent studies~\cite{LBM_ICCV2025,Resshift_NIPS2023,FlowR_ICCV25} further show that initializing from low-quality observations, rather than Gaussian noise, reduces the transport distance and improves structural fidelity by anchoring the generation process to the input.

\section{Method}

Given a noisy RGB image $\mathbf{I}_\mathrm{N}\in\mathbb{R}^{H\times W\times 3}$ and an unaligned NIR guidance image $\mathbf{I}_\mathrm{G}\in\mathbb{R}^{H\times W}$ captured by different cameras, our goal is to recover the clean RGB image $\mathbf{I}_\mathrm{T}\in\mathbb{R}^{H\times W\times 3}$. The spatial misalignment induces cross-modal displacements spanning a 2D range $[-D,D]^2$, requiring joint image restoration and correspondence estimation under severe noise.

To tackle this challenge, we propose UGD-FM, an unaligned image guided denoising framework based on cross-modal conditional flow matching. We first introduce the \textbf{conditional flow formulation} and \textbf{overall network architecture} in Sec.~\ref{subsec:overview}. We then describe the \textbf{stage-focused and inference-consistent training strategy} for effective few-step flow matching in Sec.~\ref{subsec:training_strategy}, followed by \textbf{adaptive sparse guidance propagation} for efficient guidance aggregation in Sec.~\ref{subsec:guidance_propagation}. Finally, we present the training losses in Sec.~\ref{subsec:loss}.

\subsection{Overview}
\label{subsec:overview}

\textbf{Conditional Flow Formulation.}
Previous guided denoising approaches~\cite{DVN_AAAI22, MNNet_IF22, SANet_CVPR23, RFFNet_MM24, NAID_TMM24} typically learn a direct mapping from the noisy target and guidance images to the clean target image. However, this paradigm is less effective under severe target noise and spatial misalignment, since reliable cross-modal correspondences must be estimated from a highly corrupted target state. To address this issue, we formulate unaligned guided denoising as a conditional flow matching process that progressively transports the noisy RGB distribution to the clean RGB distribution.

For each training triplet $(\mathbf{I}_\mathrm{N},\mathbf{I}_\mathrm{G},\mathbf{I}_\mathrm{T})$, we define a linear path~\cite{RectifiedFlow_Arxiv2209, FlowR_ICCV25} that satisfies
\begin{equation}
    \mathbf{z}_t=(1-t)\mathbf{I}_\mathrm{N}+t\mathbf{I}_\mathrm{T}, \quad t\in[0,1],
    \label{eq:interpolate_state}
\end{equation}
where $\mathbf{z}_0=\mathbf{I}_\mathrm{N}$ and $\mathbf{z}_1=\mathbf{I}_\mathrm{T}$. The corresponding target velocity is
\begin{equation}
    \mathbf{u}_t=\frac{d\mathbf{z}_t}{dt}=\mathbf{I}_\mathrm{T}-\mathbf{I}_\mathrm{N}.
\end{equation}
We train a conditional velocity network $v_\theta$ to predict this velocity from the current state $\mathbf{z}_t$, the unaligned guidance $\mathbf{I}_\mathrm{G}$, and the time step $t$. The training objective is
\begin{equation}
    \mathcal{L}_{\mathrm{fm}}=\mathbb{E}_{(\mathbf{I}_\mathrm{N},\mathbf{I}_\mathrm{G},\mathbf{I}_\mathrm{T}),t}
    \left\|v_\theta(\mathbf{z}_t,\mathbf{I}_\mathrm{G},t)-(\mathbf{I}_\mathrm{T}-\mathbf{I}_\mathrm{N})\right\|_1.
    \label{eq:loss_fm}
\end{equation}
During inference, we initialize $\mathbf{z}_0=\mathbf{I}_\mathrm{N}$, and solve
$d\mathbf{z}_t/dt=v_\theta(\mathbf{z}_t,\mathbf{I}_\mathrm{G},t)$ with a few-step Euler solver. Unlike direct regression, this formulation provides a progressive denoising trajectory. Along this trajectory, the target state $\mathbf{z}_t$ is encouraged to become cleaner and more structurally reliable as $t$ increases, which facilitates cross-modal correspondence estimation. The aggregated guidance further improves velocity prediction, forming a mutually reinforcing process between guidance aggregation and image restoration.

\begin{figure}[t] 
    \centering
    \includegraphics[width=\linewidth]{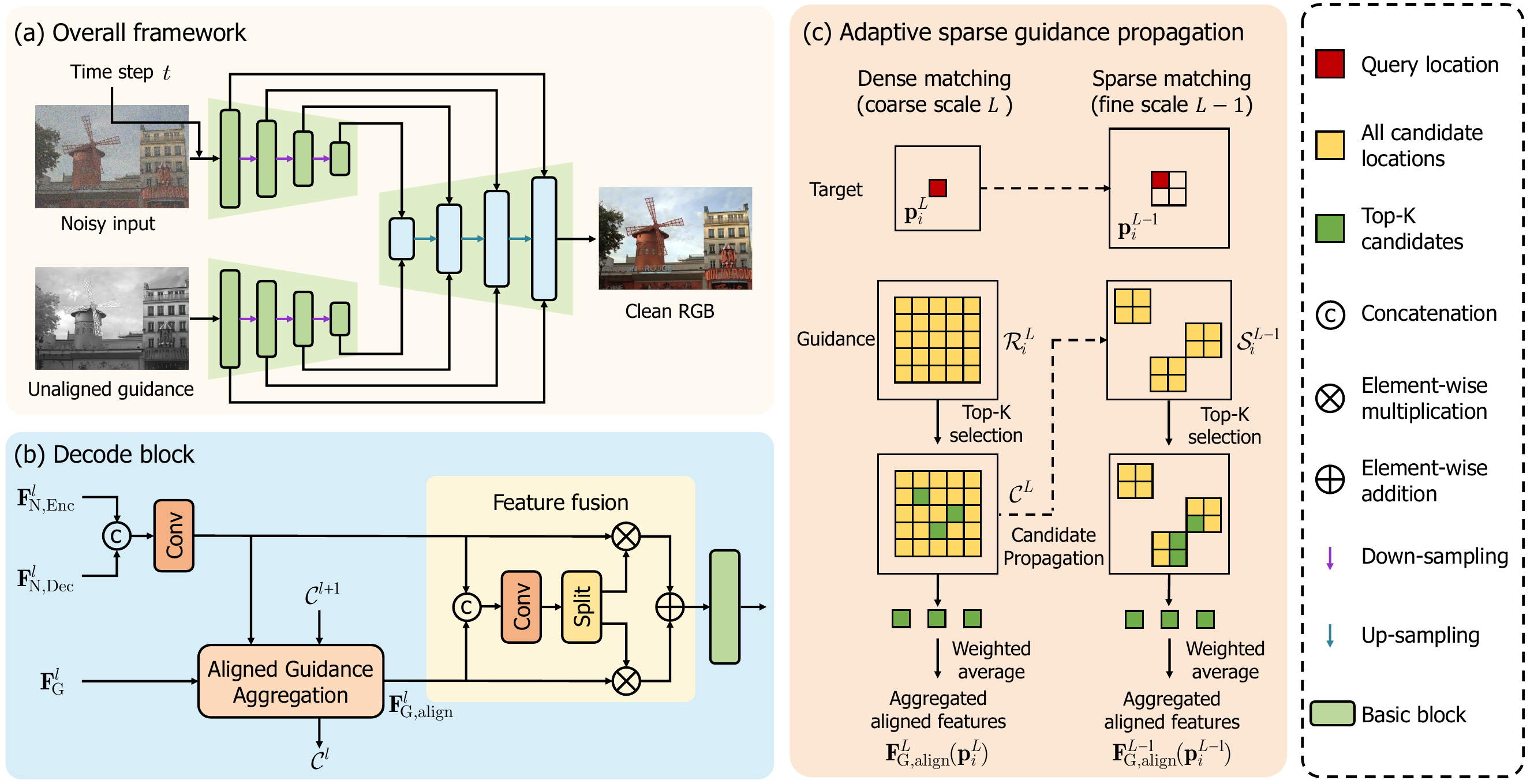} 
    \caption{Network architecture of UGD-FM. (a) UGD-FM adopts a two-branch framework for unaligned RGB-NIR guided denoising. (b) The decoder block aggregates aligned guidance features and fuses them with target features. (c) Adaptive sparse guidance propagation transfers reliable coarse candidates to finer scales for efficient sparse matching and guidance aggregation.}
    \label{fig:method}
\end{figure}

\textbf{Overall Network Architecture.} 
As illustrated in Fig.~\ref{fig:method}(a), we implement the velocity network with a two-branch U-Net~\cite{UNet_MICCAI15}. We first extract multi-scale RGB and NIR features from the current denoising state $\mathbf{z}_t$ and the guidance image $\mathbf{I}_\mathrm{G}$, respectively. The decoder then performs cross-modal guidance aggregation and feature fusion in a coarse-to-fine manner. As shown in Fig.~\ref{fig:method}(b), at each decoding scale, we first perform cross-modal guidance aggregation to obtain aligned NIR features, and then inject them into the target features through spatially variant feature fusion~\cite{SGDFormer_IF25}. Across decoding scales, we further introduce adaptive sparse guidance propagation, as shown in Fig.~\ref{fig:method}(c), to avoid redundant large-range feature matching. Following feature alignment, a gated feature fusion module is added to adaptively adjust the role of the guiding information, preventing the injection of erroneous information in case of misalignment.

We highlight that both the conditional flow formulation and the guidance aggregation design are compatible with modern single-image restoration backbones, such as NAFNet~\cite{NAFNet_ECCV22}, Uformer~\cite{Uformer_CVPR22}, and Restormer~\cite{Restormer_CVPR22}. Thus, our UGD-FM provides a general framework for adapting existing single-image denoisers to unaligned cross-modal guided denoising.

\subsection{Stage-Focused and Inference-Consistent Training}
\label{subsec:training_strategy}
Although the conditional flow formulation improves over direct regression, directly applying standard flow matching is still suboptimal for efficient image restoration. Increasing solver steps introduces additional computation, and may accumulate prediction errors along solver-induced inference trajectories. Therefore, we aim to learn an effective and efficient few-step model, \eg with one-step or two-step inference, for high-quality guided denoising.

To this end, we revisit the denoising trajectory of unaligned guided denoising, and identify two task-critical stages. The \textbf{highly corrupted stage} with small $t$ focuses on coarse RGB structure recovery, while the \textbf{relatively cleaner stage} with larger $t$ enables more reliable cross-modal matching and fine-detail restoration. Based on this observation, we place the few-step inference timesteps around these two stages, such that the solver directly visits the states most relevant to guided denoising. We then align training with this stage-focused inference schedule at two levels, \ie stage-focused time sampling aligns the training timesteps with the inference anchors, while inference-consistent state construction aligns the training states with solver-induced inference states.

\textbf{Stage-Focused Time Sampling.}
We first sample $t$ uniformly from $[0,1]$ for stable velocity learning, and then switch to stage-focused sampling to match the few-step inference schedule. Specifically, we use two inference anchors $\tau_{\mathrm{hc}}$ and $\tau_{\mathrm{rc}}$ for the highly corrupted and relatively cleaner stages, respectively, where $\tau_{\mathrm{hc}}=0$ and $0<\tau_{\mathrm{rc}}<1$. With training iteration $n$ and warm-up length $N_{\mathrm{warm}}$, the training time distribution is defined as
\begin{equation}
    p_{\mathrm{train}}^{(n)}=
        \begin{cases}
        \mathcal{U}(0,1), & n < N_{\mathrm{warm}},\\
        p_{\mathrm{sf}}, & n \geq N_{\mathrm{warm}},
        \end{cases}
\end{equation}
where $\mathcal{U}(0,1)$ denotes the uniform distribution over $[0,1]$. The stage-focused distribution $p_{\mathrm{sf}}$ is defined as a mixture of two truncated Gaussian distributions as
\begin{equation}
    p_{\mathrm{sf}}=
    \lambda_{\mathrm{hc}}q_{\mathrm{hc}} + \lambda_{\mathrm{rc}}q_{\mathrm{rc}}, \quad
    \lambda_{\mathrm{hc}}+\lambda_{\mathrm{rc}}=1,
\end{equation}
with
\begin{equation}
    q_s=\mathcal{N}_{[0,1]}(\tau_s,\sigma_s^2), \quad s\in\{\mathrm{hc},\mathrm{rc}\}.
\end{equation}
Here, $\sigma_s$ controls the sampling concentration around the anchor, and $\mathcal{N}_{[0,1]}$ denotes a Gaussian distribution truncated and normalized on $[0,1]$.

\textbf{Inference-Consistent State Construction.}
Although stage-focused sampling aligns training and inference at the timestep level, the relatively cleaner stage still has a state mismatch. It is reached after the first solver update during inference, whereas the network is trained on the interpolation state $\mathbf{z}_t$ in Eq.~\ref{eq:interpolate_state}. We therefore replace the interpolation state with a solver-induced state for samples from $q_{\mathrm{rc}}$. For $t\sim q_{\mathrm{rc}}$, we first predict the initial velocity as $\mathbf{v}_{0} = v_{\theta} \left( \mathbf{I}_{\mathrm{N}}, \mathbf{I}_{\mathrm{G}}, 0 \right)$. The inference-consistent training state is then constructed by a local Euler update as $\tilde{\mathbf{z}}_t=\mathbf{I}_{\mathrm{N}}+t\mathbf{v}_{0}$. Finally, the corresponding velocity target can be defined as the residual direction to the clean target as $\tilde{\mathbf{u}}_t = (\mathbf{I}_\mathrm{T}-\tilde{\mathbf{z}}_t)/(1-t)$.

By combining stage-focused time sampling with inference-consistent state construction, UGD-FM better aligns training with few-step inference at both the timestep and state levels, enabling high-quality guided denoising with only one or two solver steps.

\subsection{Adaptive Sparse Guidance Propagation}
\label{subsec:guidance_propagation}
Large-range cross-modal matching is essential for unaligned guided denoising. A straightforward solution is to perform dense large-range matching~\cite{SANet_CVPR23, SGDFormer_IF25, 3MTI_CVPR2026} at every decoding scale. However, it is computationally expensive, especially at high resolutions. Moreover, effective guidance aggregation mainly depends on a sparse set of reliable correspondences rather than exhaustive dense relations. We therefore propose adaptive sparse guidance propagation, which discovers reliable candidates at a coarse scale and propagates them to finer scales for sparse refinement, as illustrated in Fig.~\ref{fig:method}(c).

At scale $l$, let $s_i^l(\mathbf{p})$ denote the similarity score between target location $\mathbf{p}_i^l$ and guidance location $\mathbf{p}$, and let $\mathbf{v}^l(\mathbf{p})$ denote the guidance value feature at $\mathbf{p}$. A larger $l$ indicates a coarser scale.

\textbf{Coarse Candidate Discovery.}
At the coarsest scale $L$, we search within the range
$\mathcal{R}_i^L=\{\mathbf{p}\mid \mathbf{p}-\mathbf{p}_i^L\in[-D/2^{L-1},D/2^{L-1}]^2\}$
for each target location $\mathbf{p}_i^L$, and retain the top-$K$ guidance candidates as
\begin{equation}
    \mathcal{C}_i^{L}
    =
    \left\{
    (\mathbf{p},\alpha_i^L(\mathbf{p}))
    \mid
    \mathbf{p}\in
    \operatorname{TopK}_{\mathbf{p}\in\mathcal{R}_i^{L}}
    \left(s_i^{L}(\mathbf{p})\right)
    \right\},
\end{equation}
where $\alpha_i^L(\mathbf{p})$ is obtained by applying softmax over the selected top-$K$ candidates. The coordinates are propagated to finer scales, while the weights are used for aligned guidance aggregation. Thus, $\mathcal{C}_i^{L}$ preserves reliable large-displacement candidates, and discards most irrelevant regions.

\textbf{Sparse Candidate Propagation.}
At a finer scale $l<L$, we refine the candidates propagated from the previous coarser scale. Let $\pi(i)$ denote the parent location of target location $i$ at scale $l+1$. We expand the coordinates in $\mathcal{C}_{\pi(i)}^{l+1}$ to their $2\times2$ child locations at scale $l$, forming a sparse search set as
\begin{equation}
    \mathcal{S}_i^{l}
    =
    \operatorname{Child}
    \left(
    \{\mathbf{p}\mid(\mathbf{p},\alpha)\in\mathcal{C}_{\pi(i)}^{l+1}\}
    \right),
\end{equation}
where $\operatorname{Child}(\cdot)$ maps each coarse candidate coordinate to its $2\times2$ child locations at the current scale. We then select the top-$K$ candidates only within this sparse set as
\begin{equation}
    \mathcal{C}_i^{l}
    =
    \left\{
    (\mathbf{p},\alpha_i^l(\mathbf{p}))
    \mid
    \mathbf{p}\in
    \operatorname{TopK}_{\mathbf{p}\in\mathcal{S}_i^{l}}
    \left(s_i^{l}(\mathbf{p})\right)
    \right\},
\end{equation}
where $\alpha_i^l(\mathbf{p})$ is normalized over the selected top-$K$ candidates. The refined candidates are used for guidance aggregation at scale $l$, and further propagated to the next finer scale.

\textbf{Aligned Guidance Aggregation.}
Given $\mathcal{C}_i^{l}$, we perform aligned guidance aggregation only over the selected sparse candidates as $ \mathbf{F}_{\mathrm{G,align}}^{l}(\mathbf{p}_i^l) = \sum_{(\mathbf{p},\alpha_i^l(\mathbf{p}))\in\mathcal{C}_i^{l}} \alpha_i^l(\mathbf{p})\,\mathbf{v}^{l}(\mathbf{p})$. Following SGDFormer~\cite{SGDFormer_IF25}, we further fuse the aligned guidance and target features using spatially variant weights, which adaptively modulate each location and mitigate artifacts from imprecise matches. In this way, we avoid redundant dense matching, and aggregate only the most useful cross-modal cues for guided denoising.

\subsection{Loss Functions}
\label{subsec:loss}

Although adaptive sparse guidance propagation avoids exhaustive dense matching at finer scales, learning reliable sparse candidates remains challenging. To regularize training, we introduce two auxiliary losses, namely \textbf{perceptual consistency loss} and \textbf{flow consistency loss}, which provide auxiliary perceptual regularization and explicit displacement supervision, respectively. These losses encourage the model to select reliable guidance information for restoration.

\textbf{Perceptual Consistency Loss.}
To encourage feature-space compatibility between the aggregated NIR guidance and the RGB target, we apply the learned aggregation weights to the guidance image $\mathbf{I}_\mathrm{G}$ and obtain an aligned guidance image at scale $l$ as $ \mathbf{I}_{\mathrm{G,align}}^{l}(\mathbf{p}_i^l) = \sum_{(\mathbf{p},\alpha_i^l(\mathbf{p}))\in\mathcal{C}_i^{l}} \alpha_i^l(\mathbf{p})\,\mathbf{I}_{\mathrm{G}}^{l}(\mathbf{p})$, where $\mathbf{I}_{\mathrm{G}}^{l}$ is the downsampled NIR image at scale $l$. We then compute the perceptual consistency loss between the aligned guidance image and the clean RGB target as
\begin{equation}
    \mathcal{L}_{\mathrm{pc}}
    =
    \sum_l
    \mathcal{P}
    \left(
    \mathbf{I}_{\mathrm{G,align}}^{l},
    \mathbf{I}_{\mathrm{T}}^{l}
    \right),
\end{equation}
where $\mathcal{P}(\cdot,\cdot)$ denotes the LPIPS~\cite{LPIPS_ECCV16} loss, and $\mathbf{I}_{\mathrm{T}}^{l}$ is the clean RGB target downsampled to scale $l$.

\textbf{Flow Consistency Loss.}
To further constrain sparse correspondence learning, we supervise the displacement induced by the selected candidates with pseudo flow annotations. The pseudo annotations are precomputed by an existing flow estimation network~\cite{CrossModalRAFT_AAAI22} between the clean RGB target $\mathbf{I}_\mathrm{T}$ and the guidance image $\mathbf{I}_\mathrm{G}$, and are used as geometric supervision during training. Specifically, we compute a soft displacement field from the aggregation weights as $\mathbf{f}^{l}(\mathbf{p}_i^l) = \sum_{(\mathbf{p},\alpha_i^l(\mathbf{p}))\in\mathcal{C}_i^{l}} \alpha_i^l(\mathbf{p}) \left( \mathbf{p}-\mathbf{p}_i^l \right)$.
Let $\mathbf{f}_{\mathrm{pseudo}}^{l}$ denote the pseudo displacement field downsampled to scale $l$. The flow consistency loss is defined as
\begin{equation}
    \mathcal{L}_{\mathrm{fc}}
    =
    \sum_l \sum_i
    \left\|
    \mathbf{f}^{l}(\mathbf{p}_i^l)
    -
    \mathbf{f}_{\mathrm{pseudo}}^{l}(\mathbf{p}_i^l)
    \right\|_1.
\end{equation}
This loss provides geometric supervision for sparse candidate selection, and further encourages reliable cross-modal guidance aggregation.

\textbf{Overall Objective.}
The final training objective is
\begin{equation}
    \mathcal{L}_{\mathrm{total}} = \mathcal{L}_{\mathrm{fm}} + \lambda_{\mathrm{pc}}\mathcal{L}_{\mathrm{pc}} + \lambda_{\mathrm{fc}}\mathcal{L}_{\mathrm{fc}},
\end{equation}
where $\mathcal{L}_{\mathrm{fm}}$ is the flow matching loss in Eq.~\ref{eq:loss_fm}, and $\lambda_{\mathrm{pc}}$ and $\lambda_{\mathrm{fc}}$ balance the auxiliary losses. We note that $\mathcal{L}_{\mathrm{pc}}$ and $\mathcal{L}_{\mathrm{fc}}$ are used only during training to provide perceptual regularization and geometric supervision, respectively, and introduce no additional inference cost.

\begin{table}[t]
	\centering
	\caption{Quantitative comparison on the FANVID~\cite{FANVID_ECCV24} dataset under Gaussian noise ($\sigma=0.3$), Poisson-Gaussian noise ($\alpha=0.02, \sigma=0.2$), and Poisson-Gaussian noise ($\alpha=0.03, \sigma=0.3$). The \textbf{best} and \underline{second-best} results are highlighted in bold and underlined, respectively.}
	\label{tab:cmp_FANVID}
	\fitTableToWidth
	{
		\small
		\renewcommand{\tabcolsep}{2.5pt}{}
		\renewcommand{\arraystretch}{1.1}
		\begin{tabular}{lccccccccc}
			\toprule
			\multirow{2}{*}{\textbf{Method}} & 
			\multicolumn{3}{c}{$\sigma=0.3$} & 
			\multicolumn{3}{c}{$\alpha=0.02, \sigma=0.2$} & 
			\multicolumn{3}{c}{$\alpha=0.03, \sigma=0.3$} \\
			\cmidrule(lr){2-4} \cmidrule(lr){5-7} \cmidrule(lr){8-10}
			 & PSNR $\uparrow$ & SSIM $\uparrow$ & LPIPS $\downarrow$
			 & PSNR $\uparrow$ & SSIM $\uparrow$ & LPIPS $\downarrow$
			 & PSNR $\uparrow$ & SSIM $\uparrow$ & LPIPS $\downarrow$\\
			\midrule
			MPRNet~\cite{MPRNet_CVPR21} & 29.48 & 0.8039 & 0.3480 & 30.82 & 0.8358 & 0.3040 & 29.25 & 0.7995 & 0.3520  \\
			NAFNet~\cite{NAFNet_ECCV22} & 29.05 & 0.7932 & 0.3541 & 30.28 & 0.8239 & 0.3165 & 28.84 & 0.7891 & 0.3583  \\
			Uformer~\cite{Uformer_CVPR22} & 29.44 & 0.8037 & 0.3437 & 30.76 & 0.8358 & 0.3003 & 29.22 & 0.7995 & 0.3464  \\
			Restormer~\cite{Restormer_CVPR22} & 29.42 & 0.8024 & 0.3459 & 30.75 & 0.8339 & 0.3044 & 29.20 & 0.7981 & 0.3497  \\
			Xformer~\cite{Xformer_ICLR24} & 29.36 & 0.8008 & 0.3514 & 30.69 & 0.8332 & 0.3078 & 29.14 & 0.7967 & 0.3548  \\
			\midrule
			CUNet~\cite{CUNet_TPAMI20} & 28.16 & 0.7618 & 0.4097 & 29.55 & 0.8023 & 0.3711 & 27.87 & 0.7537 & 0.4154  \\
			DVN~\cite{DVN_AAAI22} & 29.27 & 0.7985 & 0.3592 & 30.60 & 0.8311 & 0.3152 & 29.04 & 0.7938 & 0.3631  \\
			RFFNet~\cite{RFFNet_MM24} & 29.34 & 0.8013 & 0.3506 & 30.67 & 0.8332 & 0.3074 & 29.12 & 0.7968 & 0.3546  \\
			NAID~\cite{NAID_TMM24} & 29.21 & 0.7976 & 0.3504 & 30.49 & 0.8297 & 0.3098 & 28.99 & 0.7933 & 0.3540  \\
			SANet~\cite{SANet_CVPR23} & 29.15 & 0.8072 & 0.3611 & 30.45 & 0.8388 & 0.3194 & 28.92 & 0.8029 & 0.3651  \\
			\midrule
			UGD-FM (NAFNet, 1-step) & 30.34 & 0.8384 & 0.2313 & 31.55 & 0.8648 & 0.2015 & 30.12 & 0.8351 & 0.2351  \\
			UGD-FM (NAFNet, 2-step) & 30.62 & 0.8486 & 0.2210 & 31.79 & 0.8722 & 0.1964 & 30.42 & 0.8458 & 0.2237  \\
			UGD-FM (Uformer, 1-step) & 30.32 & 0.8342 & 0.2308 & 31.54 & 0.8652 & 0.2010 & 30.09 & 0.8343 & 0.2351  \\
			UGD-FM (Uformer, 2-step) & \underline{30.66} & \underline{0.8499} & 0.2201 & \underline{31.81} & \underline{0.8732} & 0.1962 & \underline{30.45} & \underline{0.8467} & 0.2229  \\
			UGD-FM (Restormer, 1-step) & 30.50 & 0.8436 & \underline{0.2179} & 31.71 & 0.8694 & \underline{0.1909} & 30.29 & 0.8401 & \underline{0.2215}  \\
			UGD-FM (Restormer, 2-step) & \textbf{30.77} & \textbf{0.8535} & \textbf{0.2098} & \textbf{31.93} & \textbf{0.8764} & \textbf{0.1870} & \textbf{30.58} & \textbf{0.8506} & \textbf{0.2121}  \\
			\bottomrule
		\end{tabular}
	}
\end{table}

\begin{table}[t]
	\centering
	\caption{Quantitative comparison on the RGB-NIR Scene dataset~\cite{Scene_CVPR11} under Gaussian noise ($\sigma=0.3$), Poisson-Gaussian noise ($\alpha=0.02, \sigma=0.2$), and Poisson-Gaussian noise ($\alpha=0.03, \sigma=0.3$). The \textbf{best} and \underline{second-best} results are highlighted in bold and underlined, respectively.}
	\label{tab:cmp_Scene}
	\fitTableToWidth
	{
		\small
		\renewcommand{\tabcolsep}{2.5pt}{}
		\renewcommand{\arraystretch}{1.1}
		\begin{tabular}{lccccccccc}
			\toprule
			\multirow{2}{*}{\textbf{Method}} & 
			\multicolumn{3}{c}{$\sigma=0.3$} & 
			\multicolumn{3}{c}{$\alpha=0.02, \sigma=0.2$} & 
			\multicolumn{3}{c}{$\alpha=0.03, \sigma=0.3$} \\
			\cmidrule(lr){2-4} \cmidrule(lr){5-7} \cmidrule(lr){8-10}
			 & PSNR $\uparrow$ & SSIM $\uparrow$ & LPIPS $\downarrow$
			 & PSNR $\uparrow$ & SSIM $\uparrow$ & LPIPS $\downarrow$
			 & PSNR $\uparrow$ & SSIM $\uparrow$ & LPIPS $\downarrow$\\
			\midrule
			MPRNet~\cite{MPRNet_CVPR21} & 29.37 & 0.7887 & 0.3428 & 30.55 & 0.8189 & 0.3069 & 29.12 & 0.7831 & 0.3480  \\
			NAFNet~\cite{NAFNet_ECCV22} & 29.07 & 0.7851 & 0.3409 & 30.23 & 0.8152 & 0.3068 & 28.81 & 0.7794 & 0.3466  \\
			Uformer~\cite{Uformer_CVPR22} & 29.35 & 0.7890 & 0.3370 & 30.54 & 0.8201 & 0.3010 & 29.09 & 0.7833 & 0.3407  \\
			Restormer~\cite{Restormer_CVPR22} & 29.38 & 0.7897 & 0.3365 & 30.56 & 0.8198 & 0.3014 & 29.12 & 0.7841 & 0.3417  \\
			Xformer~\cite{Xformer_ICLR24} & 29.46 & 0.7912 & 0.3349 & 30.65 & 0.8216 & 0.2978 & 29.19 & 0.7859 & 0.3394  \\
			\midrule
			CUNet~\cite{CUNet_TPAMI20} & 27.37 & 0.7248 & 0.4484 & 28.70 & 0.7709 & 0.4046 & 27.00 & 0.7115 & 0.4612  \\
			DVN~\cite{DVN_AAAI22} & 29.14 & 0.7831 & 0.3528 & 30.32 & 0.8142 & 0.3168 & 28.86 & 0.7771 & 0.3587  \\
			RFFNet~\cite{RFFNet_MM24} & 29.06 & 0.7817 & 0.3540 & 30.25 & 0.8130 & 0.3175 & 28.79 & 0.7759 & 0.3603  \\
			NAID~\cite{NAID_TMM24} & 29.10 & 0.7856 & 0.3399 & 30.25 & 0.8155 & 0.3065 & 28.53 & 0.7801 & 0.3456  \\
			SANet~\cite{SANet_CVPR23} & 28.23 & 0.7540 & 0.3850 & 29.27 & 0.7878 & 0.3477 & 28.01 & 0.7477 & 0.3917  \\
			\midrule
			UGD-FM (NAFNet, 1-step) & 29.54 & 0.8077 & 0.2697 & 30.65 & 0.8346 & 0.2413 & 29.29 & 0.8024 & 0.2753  \\
			UGD-FM (NAFNet, 2-step) & 29.77 & 0.8163 & 0.2600 & 30.83 & 0.8407 & 0.2363 & 29.54 & 0.8117 & 0.2642  \\
			UGD-FM (Uformer, 1-step) & 29.56 & 0.8074 & 0.2706 & 30.70 & 0.8350 & 0.2408 & 29.29 & 0.8018 & 0.2768  \\
			UGD-FM (Uformer, 2-step) & \underline{29.84} & \underline{0.8172} & \underline{0.2598} & \underline{30.94} & \underline{0.8421} & 0.2351 & \underline{29.60} & \underline{0.8123} & \underline{0.2636}  \\
			UGD-FM (Restormer, 1-step) & 29.62 & 0.8095 & 0.2599 & 30.77 & 0.8369 & \underline{0.2322} & 29.35 & 0.8040 & 0.2651  \\
			UGD-FM (Restormer, 2-step) & \textbf{29.87} & \textbf{0.8188} & \textbf{0.2551} & \textbf{30.98} & \textbf{0.8437} & \textbf{0.2315} & \textbf{29.63} & \textbf{0.8141} & \textbf{0.2585}  \\
			\bottomrule
		\end{tabular}
	}
\end{table}

\begin{table}[t]
	\centering
	\caption{Quantitative comparison on the MS2~\cite{MS2_CVPR23} dataset under Gaussian noise ($\sigma=0.3$), Poisson-Gaussian noise ($\alpha=0.02, \sigma=0.2$), and Poisson-Gaussian noise ($\alpha=0.03, \sigma=0.3$). The \textbf{best} and \underline{second-best} results are highlighted in bold and underlined, respectively.}
	\label{tab:cmp_MS2}
	\fitTableToWidth
	{
		\small
		\renewcommand{\tabcolsep}{2.5pt}{}
		\renewcommand{\arraystretch}{1.1}
		\begin{tabular}{lccccccccc}
			\toprule
			\multirow{2}{*}{\textbf{Method}} & 
			\multicolumn{3}{c}{$\sigma=0.3$} & 
			\multicolumn{3}{c}{$\alpha=0.02, \sigma=0.2$} & 
			\multicolumn{3}{c}{$\alpha=0.03, \sigma=0.3$} \\
			\cmidrule(lr){2-4} \cmidrule(lr){5-7} \cmidrule(lr){8-10}
			 & PSNR $\uparrow$ & SSIM $\uparrow$ & LPIPS $\downarrow$
			 & PSNR $\uparrow$ & SSIM $\uparrow$ & LPIPS $\downarrow$
			 & PSNR $\uparrow$ & SSIM $\uparrow$ & LPIPS $\downarrow$\\
			\midrule
			MPRNet~\cite{MPRNet_CVPR21} & 30.89 & 0.8836 & 0.3201 & 32.53 & 0.9091 & 0.2785 & 30.57 & 0.8797 & 0.3253  \\
			NAFNet~\cite{NAFNet_ECCV22} & 30.76 & 0.8838 & 0.3095 & 32.36 & 0.9087 & 0.2718 & 30.48 & 0.8803 & 0.3133  \\
			Uformer~\cite{Uformer_CVPR22} & 30.10 & 0.8837 & 0.3188 & 31.42 & 0.9095 & 0.2799 & 29.84 & 0.8799 & 0.3234  \\
			Restormer~\cite{Restormer_CVPR22} & 30.98 & 0.8857 & 0.3098 & 32.62 & 0.9106 & 0.2707 & 30.69 & 0.8821 & 0.3145  \\
			Xformer~\cite{Xformer_ICLR24} & 30.34 & 0.8623 & 0.2966 & 31.98 & 0.8930 & 0.2538 & 30.01 & 0.8573 & 0.3033  \\
			\midrule
			CUNet~\cite{CUNet_TPAMI20} & 29.61 & 0.8542 & 0.3887 & 31.27 & 0.8877 & 0.3444 & 29.29 & 0.8487 & 0.3959  \\
			DVN~\cite{DVN_AAAI22} & 31.78 & 0.9096 & 0.2552 & 33.29 & 0.9272 & 0.2283 & 31.64 & 0.9077 & 0.2574  \\
			RFFNet~\cite{RFFNet_MM24} & 30.08 & 0.8685 & 0.3666 & 31.78 & 0.8971 & 0.3219 & 29.77 & 0.8643 & 0.3728  \\
			NAID~\cite{NAID_TMM24} & 30.33 & 0.8732 & 0.3362 & 31.91 & 0.9000 & 0.2965 & 30.04 & 0.8693 & 0.3412  \\
			SANet~\cite{SANet_CVPR23} & 29.68 & 0.8466 & 0.3489 & 31.27 & 0.8806 & 0.3021 & 29.33 & 0.8402 & 0.3579  \\
			\midrule
			UGD-FM (NAFNet, 1-step) & 32.42 & 0.9237 & 0.1752 & 33.79 & 0.9379 & 0.1550 & 32.14 & 0.9219 & 0.1779  \\
			UGD-FM (NAFNet, 2-step) & 32.64 & 0.9285 & 0.1685 & 33.96 & 0.9409 & 0.1524 & 32.39 & 0.9270 & 0.1704  \\
			UGD-FM (Uformer, 1-step) & 32.65 & 0.9282 & 0.1794 & 33.99 & 0.9411 & 0.1589 & 32.36 & 0.9261 & 0.1827  \\
			UGD-FM (Uformer, 2-step) & \underline{32.97} & \underline{0.9330} & 0.1648 & \underline{34.24} & \underline{0.9441} & 0.1490 & \underline{32.71} & \underline{0.9313} & 0.1667  \\
			UGD-FM (Restormer, 1-step) & 32.85 & 0.9303 & \underline{0.1635} & 34.18 & 0.9425 & \underline{0.1458} & 32.59 & 0.9285 & \underline{0.1658}  \\
			UGD-FM (Restormer, 2-step) & \textbf{33.07} & \textbf{0.9343} & \textbf{0.1592} & \textbf{34.34} & \textbf{0.9452} & \textbf{0.1444} & \textbf{32.82} & \textbf{0.9329} & \textbf{0.1609}  \\
			\bottomrule
		\end{tabular}
	}
\end{table}

\section{Experiments}

\subsection{Experimental Setting}
\textbf{Datasets.} We evaluate UGD-FM on FANVID~\cite{FANVID_ECCV24}, RGB-NIR Scene~\cite{Scene_CVPR11}, and MS2~\cite{MS2_CVPR23}. Dataset details are given in Appendix~\ref{Details_of_datasets}.

\textbf{Implementation Details.}
We implement UGD-FM with the basic blocks of NAFNet~\cite{NAFNet_ECCV22}, Uformer~\cite{Uformer_CVPR22}, and Restormer~\cite{Restormer_CVPR22}, yielding three model variants. The displacement search range is $[-D,D]^2$ with radius $D=80$. During training, degraded observations are synthesized by randomly adding Poisson-Gaussian noise to clean RGB images. Specifically, the Poisson noise parameter $\alpha$ is sampled from $[0,0.03]$, and the Gaussian noise level $\sigma$ is sampled from $[0,0.3]$. More implementation and training details are provided in Appendix~\ref{Additional_Experimental_Details}.

\textbf{Metrics.} We quantitatively evaluate the denoising performance using peak signal-to-noise ratio (PSNR), structural similarity (SSIM)~\cite{SSIM_TIP04}, and learned perceptual image patch similarity (LPIPS)~\cite{LPIPS_ECCV16}. Higher PSNR and SSIM, and lower LPIPS, indicate better performance.

\subsection{Comparisons with Existing Approaches}
We compare UGD-FM with ten representative methods, including five single-image denoisers, \ie MPRNet~\cite{MPRNet_CVPR21}, NAFNet~\cite{NAFNet_ECCV22}, Uformer~\cite{Uformer_CVPR22}, Restormer~\cite{Restormer_CVPR22}, and Xformer~\cite{Xformer_ICLR24}, and five guided denoising methods, \ie CUNet~\cite{CUNet_TPAMI20}, DVN~\cite{DVN_AAAI22}, RFFNet~\cite{RFFNet_MM24}, NAID~\cite{NAID_TMM24}, and SANet~\cite{SANet_CVPR23}. We evaluate UGD-FM with three backbone blocks, including NAFNet, Uformer, and Restormer. The denoising performance is evaluated under three noise settings: Gaussian noise with $\sigma=0.3$, Poisson-Gaussian noise with $\alpha=0.02,\sigma=0.2$, and Poisson-Gaussian noise with $\alpha=0.03,\sigma=0.3$.

\begin{figure}[t]
    \centering
    \includegraphics[width=\linewidth]
    {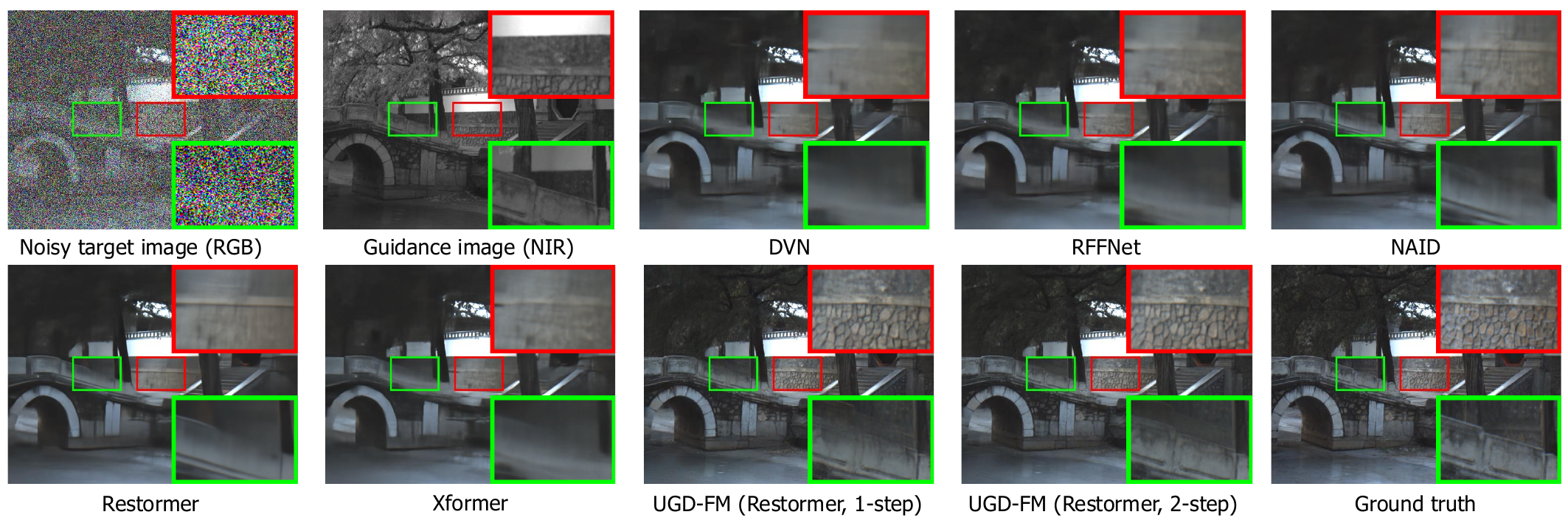}
    \includegraphics[width=\linewidth]
    {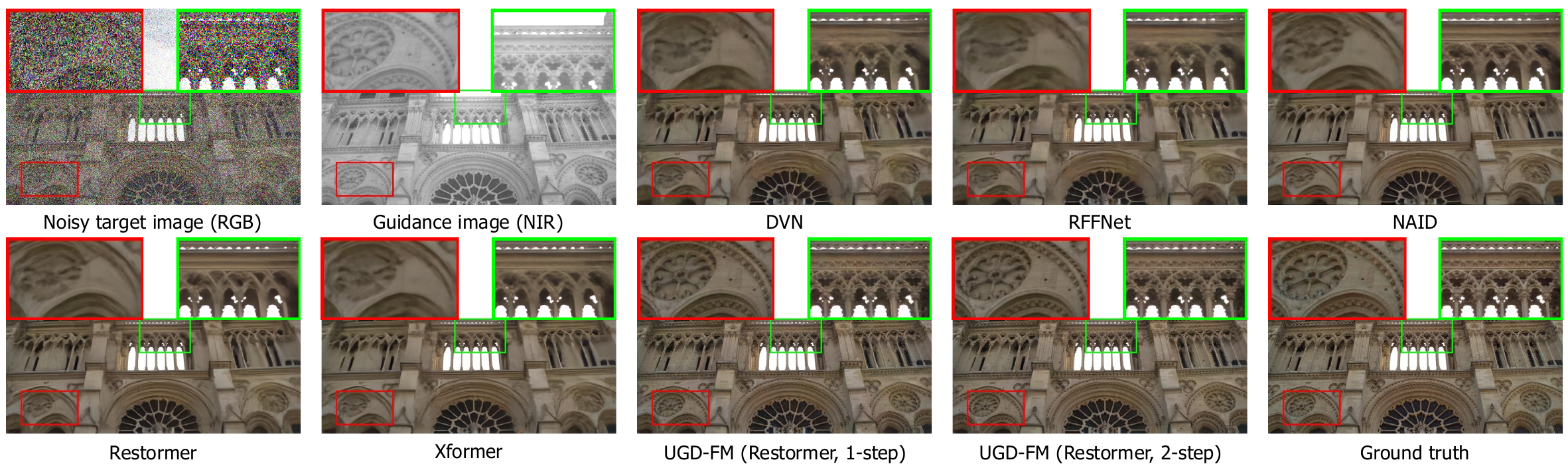}
    \includegraphics[width=\linewidth]
    {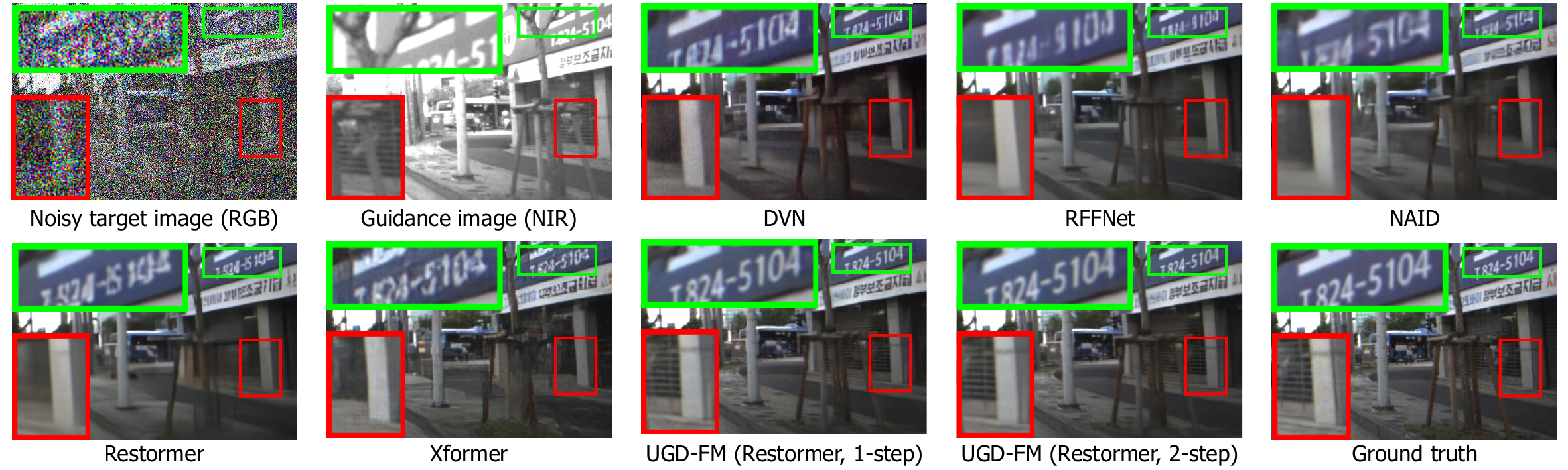}
    \caption{Qualitative comparison of image denoising results across different approaches.}
    \label{fig:vis_cmp}
\end{figure}

Tables~\ref{tab:cmp_FANVID}, \ref{tab:cmp_Scene}, and \ref{tab:cmp_MS2} report quantitative results on FANVID~\cite{FANVID_ECCV24}, RGB-NIR Scene~\cite{Scene_CVPR11}, and MS2~\cite{MS2_CVPR23}, respectively. UGD-FM consistently outperforms both single-image and guided denoising baselines across all datasets and noise settings. The gains over single-image denoisers demonstrate the benefit of exploiting NIR guidance, while the improvements over guided denoising methods show the effectiveness of UGD-FM under cross-modal misalignment. Among different variants, two-step inference further improves over one-step inference, and UGD-FM with the Restormer block achieves the best overall performance. Fig.~\ref{fig:vis_cmp} presents qualitative comparisons of different approaches. Single-image denoising methods tend to oversmooth fine structures, while existing guided methods struggle to handle unaligned cross-modal inputs. In contrast, UGD-FM better exploits unaligned NIR guidance and recovers sharper textures and more faithful structural details.

\textbf{Diffusion and Flow Matching Baselines.} We further compare our UGD-FM with more diffusion and flow matching baselines like RDBM~\cite{RDBM_CVPR26}, 3M-TI~\cite{3MTI_CVPR2026} and RFfusion~\cite{RFfusion_NIPS2025}. As shown in Table~\ref{tab:cmp_generative_fanvid}, UGD-FM outperforms these baselines in PSNR, SSIM, and LPIPS.

\textbf{RGB-Guided Thermal Image Super-Resolution.} We extend UGD-FM to RGB-guided thermal image super-resolution following the 3M-TI protocol~\cite{3MTI_CVPR2026}, and list results in Table~\ref{tab:cmp_thermal_sr}. UGD-FM achieves the best SSIM and MUSIQ, while obtaining competitive PSNR and perceptual quality, validating the generalizability and extensibility of our framework beyond guided denoising.

\begin{table}[t]
	\centering
	\begin{minipage}[t]{0.49\textwidth}
		\centering
		\captionof{table}{Comparison with more diffusion and flow matching methods on FANVID~\cite{FANVID_ECCV24} under Poisson-Gaussian noise ($\alpha=0.03$, $\sigma=0.3$). The \textbf{best} and \underline{second-best} results are highlighted in bold and underlined, respectively.}
        \vspace{1mm}
		\label{tab:cmp_generative_fanvid}
		\fitTableToWidth
		{
			\small
			\renewcommand{\tabcolsep}{2.5pt}{}
			\renewcommand{\arraystretch}{1.1}
			\begin{tabular}{lccc}
				\toprule
				Method & PSNR $\uparrow$ & SSIM $\uparrow$ & LPIPS $\downarrow$ \\
				\midrule
				RDBM~\cite{RDBM_CVPR26} & 28.80 & 0.7828 & 0.3535 \\
				3M-TI~\cite{3MTI_CVPR2026} & 28.89 & 0.7872 & 0.3221 \\
				RFfusion~\cite{RFfusion_NIPS2025} & 28.53 & 0.7702 & 0.3251 \\
				\midrule
				UGD-FM (NAFNet, 1-step) & \underline{30.12} & \underline{0.8351} & \underline{0.2351} \\
				UGD-FM (NAFNet, 2-step) & \textbf{30.42} & \textbf{0.8458} & \textbf{0.2237} \\
				\bottomrule
			\end{tabular}
		}
	\end{minipage}%
	\hfill
	\begin{minipage}[t]{0.49\textwidth}
		\centering
		\captionof{table}{Comparison of RGB-guided thermal image super-resolution following the 3M-TI~\cite{3MTI_CVPR2026} protocol. The \textbf{best} and \underline{second-best} results are highlighted in bold and underlined, respectively.}
        \vspace{1mm}
		\label{tab:cmp_thermal_sr}
		\fitTableToWidth
		{
			\small
			\renewcommand{\tabcolsep}{2.5pt}{}
			\renewcommand{\arraystretch}{1.1}
			\begin{tabular}{lccccc}
				\toprule
				Method & PSNR $\uparrow$ & SSIM $\uparrow$ & LPIPS $\downarrow$ & MANIQA $\uparrow$ & MUSIQ $\uparrow$ \\
				\midrule
				CoRPLE~\cite{CoRPLE_ECCV24} & \textbf{30.47} & \underline{0.8642} & 0.3206 & 0.2833 & 30.46 \\
				SeeSR~\cite{SeeSR_CVPR24} & 29.41 & 0.8495 & \underline{0.1828} & \underline{0.4278} & 35.22 \\
				3M-TI~\cite{3MTI_CVPR2026} & 30.09 & 0.8610 & \textbf{0.1787} & \textbf{0.4443} & \underline{36.66} \\
				UGD-FM & \underline{30.36} & \textbf{0.8710} & 0.2012 & 0.4008 & \textbf{36.94} \\
				\bottomrule
			\end{tabular}
		}
	\end{minipage}
\end{table}

\subsection{Ablation Study}

\begin{table}[t]
	\centering
	\caption{Ablation studies on FANVID~\cite{FANVID_ECCV24} under Poisson-Gaussian noise ($\alpha=0.03$, $\sigma=0.3$).}
	\label{tab_ablation}
	\begin{subtable}{\linewidth}
		\centering
		\caption{Flow matching formulation and training strategy.}
		\label{tab:ablation_training}
		\fitTableToWidth
		{
			\small
			\renewcommand{\tabcolsep}{4pt}{}
			\renewcommand{\arraystretch}{1.1}
			\begin{tabular}{lcccccc}
				\toprule
				\multirow{2}{*}{Setting} & \multicolumn{3}{c}{1-step} & \multicolumn{3}{c}{2-step} \\
				\cmidrule(lr){2-4} \cmidrule(lr){5-7}
				 & PSNR $\uparrow$ & SSIM $\uparrow$ & LPIPS $\downarrow$ & PSNR $\uparrow$ & SSIM $\uparrow$ & LPIPS $\downarrow$ \\
				\midrule
				Direct regression & 30.27 & 0.8367 & 0.2376 & -- & -- & -- \\
				Standard FM & 30.28 & 0.8389 & 0.2357 & 30.34 & 0.8401 & 0.2312 \\
				\quad + Stage-focused sampling & \textbf{30.31} & 0.8396 & 0.2238 & 30.42 & 0.8421 & 0.2224 \\
				\quad + Inference-consistent state construction & 30.29 & \textbf{0.8401} & \textbf{0.2215} & \textbf{30.58} & \textbf{0.8506} & \textbf{0.2121} \\
				\bottomrule
			\end{tabular}
		}
	\end{subtable}

	\begin{subtable}{\linewidth}
		\centering
		\caption{Guidance aggregation.}
		\label{tab:ablation_aggregation}
		\fitTableToWidth
		{
			\small
			\renewcommand{\tabcolsep}{4pt}{}
			\renewcommand{\arraystretch}{1.1}
			\begin{tabular}{lcccccc}
				\toprule
				\multirow{2}{*}{Setting} & \multicolumn{3}{c}{1-step} & \multicolumn{3}{c}{2-step} \\
				\cmidrule(lr){2-4} \cmidrule(lr){5-7}
				 & PSNR $\uparrow$ & SSIM $\uparrow$ & LPIPS $\downarrow$ & PSNR $\uparrow$ & SSIM $\uparrow$ & LPIPS $\downarrow$ \\
				\midrule
				No attention & 28.97 & 0.7920 & 0.3247 & 29.10 & 0.7976 & 0.3244 \\
                Global attention & OOM & OOM & OOM & OOM & OOM & OOM \\
				Window attention & 29.64 & 0.8188 & 0.2512 & 30.03 & 0.8370 & 0.2369 \\
				Sparse propagation & \textbf{30.29} & \textbf{0.8401} & \textbf{0.2215} & \textbf{30.58} & \textbf{0.8506} & \textbf{0.2121} \\
				\bottomrule
			\end{tabular}
		}
	\end{subtable}

	\begin{subtable}{\linewidth}
		\centering
		\caption{Auxiliary losses.}
		\label{tab:ablation_losses}
		\fitTableToWidth
		{
			\small
			\renewcommand{\tabcolsep}{4pt}{}
			\renewcommand{\arraystretch}{1.1}
			\begin{tabular}{cccccccc}
				\toprule
				\multirow{2}{*}{$\mathcal{L}_{\mathrm{pc}}$} & \multirow{2}{*}{$\mathcal{L}_{\mathrm{fc}}$} & \multicolumn{3}{c}{1-step} & \multicolumn{3}{c}{2-step} \\
				\cmidrule(lr){3-5} \cmidrule(lr){6-8}
				 & & PSNR $\uparrow$ & SSIM $\uparrow$ & LPIPS $\downarrow$ & PSNR $\uparrow$ & SSIM $\uparrow$ & LPIPS $\downarrow$ \\
				\midrule
				\xmark & \xmark & 29.48 & 0.8062 & 0.2862 & 29.78 & 0.8213 & 0.2660 \\
				\xmark & \cmark & 30.13 & 0.8355 & 0.2306 & 30.45 & 0.8465 & 0.2217 \\
				\cmark & \xmark & 30.01 & 0.8324 & 0.2345 & 30.27 & 0.8445 & 0.2234 \\
				\cmark & \cmark & \textbf{30.29} & \textbf{0.8401} & \textbf{0.2215} & \textbf{30.58} & \textbf{0.8506} & \textbf{0.2121} \\
				\bottomrule
			\end{tabular}
		}
	\end{subtable}
\end{table}

We conduct ablation studies on FANVID~\cite{FANVID_ECCV24} under Poisson-Gaussian noise with $\alpha=0.03$ and $\sigma=0.3$. We use UGD-FM with the Restormer block as the default setting, and list results in Table~\ref{tab_ablation}.

\textbf{Flow Matching Formulation and Training Strategy.} We first examine the effect of the flow formulation. As shown in Table~\ref{tab:ablation_training}, compared with direct regression, base flow matching achieves comparable or slightly better one-step performance. More importantly, the flow formulation naturally supports multi-step inference, improving PSNR from $30.28$ to $30.34$. This demonstrates the benefit of modeling guided denoising as a progressive transport process. Cumulatively adding stage-focused sampling and inference-consistent states increases two-step PSNR to $30.42$ and $30.58$, respectively. Stage-focused sampling aligns the training-time distribution with the fixed inference anchors. Inference-consistent state construction further aligns the training and inference states by replacing the interpolation state with a solver-induced state.

\textbf{Adaptive Sparse Guidance Propagation.} We evaluate the efficiency of the guidance aggregation design in Table~\ref{tab:ablation_aggregation}. Without attention, the model cannot handle spatial misalignment, leading to a large performance drop. Global attention leads to out-of-memory errors, showing that dense large-range matching is impractical at high resolutions. Window attention provides a non-local receptive field but is less flexible. In contrast, our sparse propagation avoids exhaustive matching while retaining reliable candidates for guidance aggregation, enabling efficient inference.

\textbf{Auxiliary Losses.} Table~\ref{tab:ablation_losses} shows complementary benefits from perceptual regularization and displacement supervision. Removing either $\mathcal{L}_{\mathrm{pc}}$ or $\mathcal{L}_{\mathrm{fc}}$ degrades the restoration quality, and removing both losses causes a clear performance drop, with the two-step PSNR decreasing from $30.58$ to $29.78$. This suggests that perceptual and geometric regularization are complementary and important for stabilizing sparse correspondence learning and improving aligned guidance aggregation.

\section{Conclusions}
We have presented UGD-FM, a general cross-modal conditional flow matching framework for unaligned RGB-NIR guided denoising. To support efficient few-step inference, we design a stage-focused and inference-consistent training strategy that aligns training with task-critical solver states. We also introduce adaptive sparse guidance propagation with perceptual and geometric auxiliary losses for efficient and reliable large-range cross-modal matching. Extensive experiments demonstrate that UGD-FM achieves state-of-the-art denoising performance with only one or two inference steps.

\begin{ack}
This work was supported by the National Natural Science Foundation of China under Grant 62571478.
\end{ack}

\clearpage
\bibliography{references}
\bibliographystyle{ieee_fullname}
\clearpage

\appendix
\section*{Appendix}

\section{Additional Experimental Details}
\subsection{Details of Datasets}
\label{Details_of_datasets}
We evaluate UGD-FM on the RGB-NIR Scene~\cite{Scene_CVPR11}, FANVID~\cite{FANVID_ECCV24}, and MS2~\cite{MS2_CVPR23} datasets. A statistical summary of these datasets is provided in Table~\ref{tab:dataset_summary}. Specifically, FANVID contains 453 RGB images and 3624 RGB-NIR pairs with an original image resolution of $576\times768$ ($H\times W$), while RGB-NIR Scene provides 434 RGB images and 3472 RGB-NIR pairs with original resolutions ranging from $620\times502$ to $768\times1024$. MS2 consists of 1229 RGB-NIR pairs captured from real-world multi-camera imaging systems with an original resolution of $320\times1022$. We use 3008/616, 3112/360, and 981/248 image pairs for training/testing on FANVID, RGB-NIR Scene, and MS2, respectively.

RGB-NIR Scene and FANVID originally provide spatially aligned RGB-NIR image pairs, whereas MS2 is captured using real multi-camera imaging systems and naturally contains viewpoint discrepancies between modalities. Therefore, for RGB-NIR Scene and FANVID, we further construct misaligned RGB-NIR pairs through a controlled parallax simulation strategy.

\begin{figure}[t] 
    \centering
    \includegraphics[width=\linewidth]{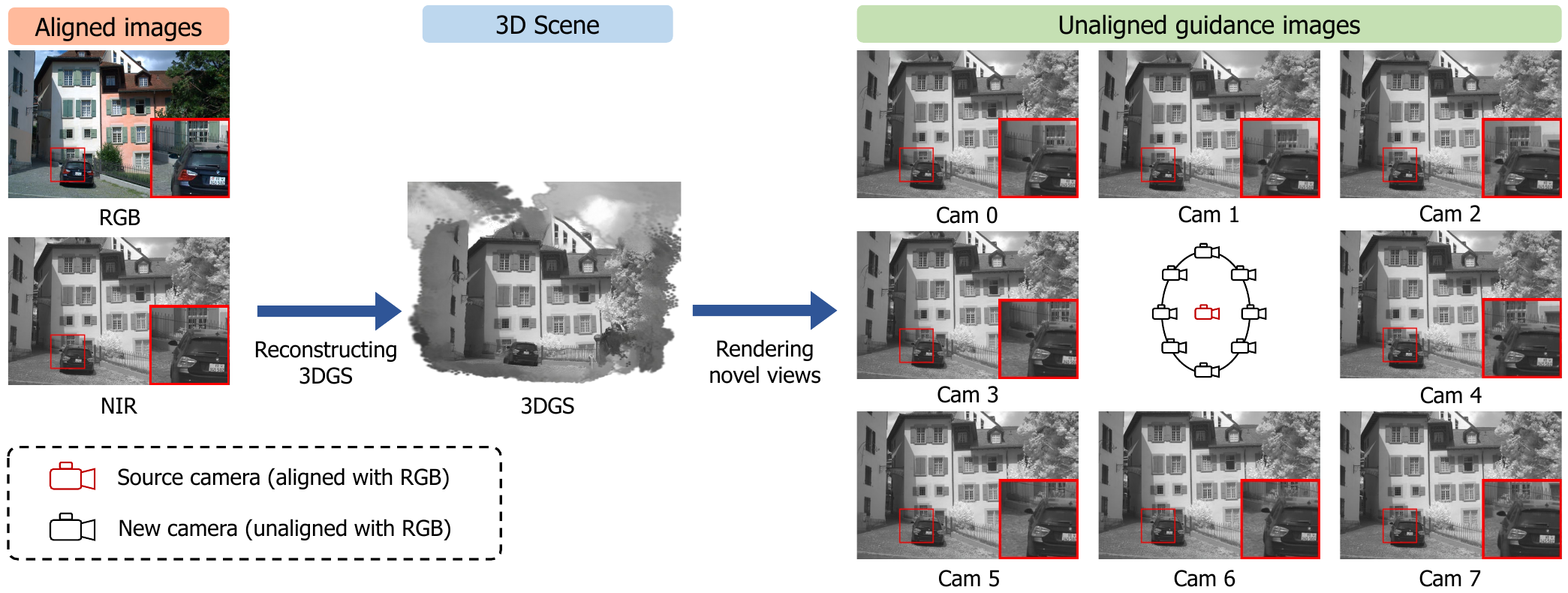} 
    \caption{Visualization of the simulation process for constructing unaligned RGB-NIR datasets.}
    \label{fig:dataset}
\end{figure}

Specifically, starting from the aligned RGB-NIR pairs, the RGB image is retained as the reference target view, while the NIR image is independently reconstructed into a 3D Gaussian representation using SHARP~\cite{ML-Sharp_ICLR26}. Based on the reconstructed representation, eight neighboring virtual camera viewpoints are sampled around the original camera pose to render new NIR views, as illustrated in Fig.~\ref{fig:dataset}. We place the virtual cameras on a circular trajectory whose radius is normalized so that the nominal translational parallax at a representative near-scene depth is approximately $x\%$ of the image diagonal. We use $x=6$ for FANVID and $x=8$ for RGB-NIR Scene. This simulation protocol is designed to approximate practical dual-camera acquisition settings, where RGB and NIR sensors observe the same scene from different viewpoints. The rendered NIR images are subsequently paired with the original RGB images to form misaligned RGB-NIR image pairs with realistic 3D parallax.

Table~\ref{tab:dataset_summary} summarizes 2D displacement magnitudes. For the synthetic FANVID and RGB-NIR Scene datasets, we obtain ground-truth flow from 3DGS rendering. For the real-world MS2 dataset, we estimate flow using CrossModalRAFT~\cite{CrossModalRAFT_AAAI22}. The P95 displacements are 45.25, 76.25, and 50.75 pixels, respectively.

\begin{table}[htbp]
	\centering
	\caption{Dataset characteristics and statistics of 2D displacement magnitudes.}
	\label{tab:dataset_summary}
	\fitTableToWidth
	{
		\small
		\renewcommand{\tabcolsep}{4pt}{}
		\renewcommand{\arraystretch}{1.1}
		\begin{tabular}{lccc}
			\toprule
			Characteristic & FANVID~\cite{FANVID_ECCV24} & RGB-NIR Scene~\cite{Scene_CVPR11} & MS2~\cite{MS2_CVPR23} \\
			\midrule
			\# RGB images & 453 & 434 & 1229 \\
			\# RGB-NIR pairs & 3624 & 3472 & 1229 \\
			Resolution ($H\times W$) & $576\times768$ & $620\times502$--$768\times1024$ & $320\times1022$ \\
			\# Training pairs & 3008 & 3112 & 981 \\
			\# Testing pairs & 616 & 360 & 248 \\
			\midrule[0.3pt]
			\addlinespace[2pt]
			Mean displacement (px) & 22.10 & 36.70 & 29.64 \\
			Std. displacement (px) & 13.46 & 23.16 & 12.28 \\
			P75 displacement (px) & 32.25 & 54.25 & 36.25 \\
			P90 displacement (px) & 41.25 & 69.25 & 43.75 \\
			P95 displacement (px) & 45.25 & 76.25 & 50.75 \\
			\bottomrule
		\end{tabular}
	}
\end{table}

\subsection{Implementation and Experimental Details}
\label{Additional_Experimental_Details}

Different variants instantiate Uformer, Restormer, or NAFNet as backbone blocks while sharing the same overall architecture. The base channel width is set to $C=32$, and only a single block is employed at each scale. The maximum network depth is set to $L=4$. The Top-$K$ parameter is fixed to $K=5$. Inference under the flow matching framework is performed using two steps at $t=\tau_{\mathrm{hc}}$ and $t=\tau_{\mathrm{rc}}$, where $\tau_{\mathrm{hc}}=0$ and $\tau_{\mathrm{rc}}=0.9$.

We optimize the network using the AdamW optimizer with a weight decay of $0.01$, an initial learning rate of $3\times10^{-4}$, and a OneCycle learning rate scheduler. Training is performed for $80\mathrm{k}$ iterations using a batch size of $8$. The warm-up stage is conducted for $N_{\mathrm{warm}}=50\mathrm{k}$ iterations. The auxiliary loss weights are set to $\lambda_{\mathrm{pc}}=0.05$ and $\lambda_{\mathrm{fc}}=0.05$, while the truncated Gaussian distribution weights are set to $\lambda_{\mathrm{hc}}=0.5$ and $\lambda_{\mathrm{rc}}=0.5$. In addition, the parameter $\sigma_s$ is set to $0.05$. 

For fair comparison, all approaches are trained using the same training set, and the training iterations are also the same. All experiments are conducted on NVIDIA RTX 4090 GPUs.

\section{More Quantitative and Qualitative Results}

\subsection{Ablation on Hyperparameters}

In Table~\ref{tab:ablation_hyperparameter}, we provide detailed ablations on hyperparameters including the displacement search radius $D$, Top-$K$ candidate selection, inference time anchors $\tau_{\mathrm{hc}}$ and $\tau_{\mathrm{rc}}$, time sampling standard deviation $\sigma_s$, and warm-up length $N_{\mathrm{warm}}$.

\begin{table}[t]
	\centering
	\caption{Hyperparameter ablations on FANVID~\cite{FANVID_ECCV24} under Poisson-Gaussian noise ($\alpha=0.03$, $\sigma=0.3$).}
	\label{tab:ablation_hyperparameter}
	\fitTableToWidth
	{
		\small
		\renewcommand{\tabcolsep}{4pt}{}
		\renewcommand{\arraystretch}{1.1}
		\begin{tabular}{cccccccc}
			\toprule
			\multirow{2}{*}{Hyperparameter} & \multirow{2}{*}{Setting} & \multicolumn{3}{c}{1-step} & \multicolumn{3}{c}{2-step} \\
			\cmidrule(lr){3-5}\cmidrule(lr){6-8}
			 &  & PSNR $\uparrow$ & SSIM $\uparrow$ & LPIPS $\downarrow$ & PSNR $\uparrow$ & SSIM $\uparrow$ & LPIPS $\downarrow$ \\
			\midrule[0.3pt]
			\addlinespace[2pt]
			\multirow{5}{*}{Search range $D$} & 32 & 30.19 & 0.8377 & 0.2241 & 30.42 & 0.8424 & 0.2179 \\
			 & 48 & 30.29 & 0.8396 & 0.2167 & 30.49 & 0.8471 & 0.2105 \\
			 & 64 & 30.32 & 0.8413 & 0.2179 & 30.56 & 0.8527 & 0.2097 \\
			 & 80 (default) & 30.29 & 0.8401 & 0.2215 & 30.58 & 0.8506 & 0.2121 \\
			 & 96 & 30.31 & 0.8403 & 0.2221 & 30.61 & 0.8507 & 0.2128 \\
			\midrule[0.3pt]
			\addlinespace[2pt]
			\multirow{4}{*}{Top-$K$ selection} & 1 & 29.89 & 0.8248 & 0.2449 & 30.19 & 0.8364 & 0.2345 \\
			 & 3 & 30.26 & 0.8382 & 0.2264 & 30.55 & 0.8492 & 0.2159 \\
			 & 5 (default) & 30.29 & 0.8401 & 0.2215 & 30.58 & 0.8506 & 0.2121 \\
			 & 7 & 30.30 & 0.8417 & 0.2185 & 30.59 & 0.8520 & 0.2110 \\
			\midrule[0.3pt]
			\addlinespace[2pt]
			\multirow{2}{*}{Time anchor $\tau_{\mathrm{hc}}$} & 0 (default) & 30.29 & 0.8401 & 0.2215 & 30.58 & 0.8506 & 0.2121 \\
			 & 0.1 & 29.85 & 0.8266 & 0.2509 & 30.21 & 0.8391 & 0.2358 \\
			\midrule[0.3pt]
			\addlinespace[2pt]
			\multirow{4}{*}{Time anchor $\tau_{\mathrm{rc}}$} & 0.3 & 30.34 & 0.8432 & 0.2235 & 30.39 & 0.8457 & 0.2216 \\
			 & 0.5 & 30.33 & 0.8412 & 0.2192 & 30.49 & 0.8459 & 0.2191 \\
			 & 0.7 & 30.33 & 0.8409 & 0.2237 & 30.56 & 0.8495 & 0.2133 \\
			 & 0.9 (default) & 30.29 & 0.8401 & 0.2215 & 30.58 & 0.8506 & 0.2121 \\
			\midrule[0.3pt]
			\addlinespace[2pt]
			\multirow{3}{*}{Sampling std. $\sigma_s$} & 0.025 & 30.33 & 0.8414 & 0.2220 & 30.62 & 0.8524 & 0.2139 \\
			 & 0.05 (default) & 30.29 & 0.8401 & 0.2215 & 30.58 & 0.8506 & 0.2121 \\
			 & 0.2 & 30.27 & 0.8384 & 0.2257 & 30.46 & 0.8453 & 0.2192 \\
			\midrule[0.3pt]
			\addlinespace[2pt]
			\multirow{3}{*}{Warm-up $N_{\mathrm{warm}}$} & $40\mathrm{k}$ & 30.31 & 0.8406 & 0.2211 & 30.56 & 0.8512 & 0.2110 \\
			 & $50\mathrm{k}$ (default) & 30.29 & 0.8401 & 0.2215 & 30.58 & 0.8506 & 0.2121 \\
			 & $60\mathrm{k}$ & 30.29 & 0.8392 & 0.2224 & 30.55 & 0.8497 & 0.2123 \\
			\bottomrule
		\end{tabular}
	}
\end{table}

\textbf{Displacement Search Radius $D$.}
We vary the displacement search radius from $32$ to $96$. A small search range may fail to cover the true displacement range, causing a performance drop, while increasing $D$ generally improves PSNR by providing a broader search range.

\textbf{Top-$K$ Candidate Selection.}
We vary $K$ from $1$ to $7$. A small $K$ may discard plausible correspondences, causing a clear performance drop. Increasing $K$ improves performance, with only a small additional gain from $5$ to $7$. We use $K=5$ as the default.

\textbf{Inference Time Anchors $\tau_{\mathrm{hc}}$ and $\tau_{\mathrm{rc}}$.}
Moving $\tau_{\mathrm{hc}}$ from $0$ to $0.1$ reduces performance, supporting the use of the initial noisy state as the highly corrupted-stage anchor. For $\tau_{\mathrm{rc}}$, an early anchor provides limited separation between the highly corrupted and relatively cleaner stages, resulting in smaller two-step gains. When the anchor clearly separates these stages, as with $\tau_{\mathrm{rc}}=0.7$ or $0.9$, two-step inference yields clear improvements. Their comparable results indicate that the improvement does not require a precisely tuned anchor. We use $\tau_{\mathrm{rc}}=0.9$ as the default.

\textbf{Time Sampling Standard Deviation $\sigma_s$.}
The parameter $\sigma_s$ controls the sampling concentration around the time anchors. Similar results at $0.025$ and $0.05$ show that the method does not require careful tuning within this range. However, an excessively large value such as $0.2$ disperses the training distribution away from the anchors and degrades performance.

\textbf{Warm-up Length $N_{\mathrm{warm}}$.}
We vary the warm-up length from $40\mathrm{k}$ to $60\mathrm{k}$ iterations. The results remain similar across these settings, indicating limited sensitivity to the warm-up length within the tested range. We use $N_{\mathrm{warm}}=50\mathrm{k}$ as the default.

\subsection{Sensitivity to Flow-Label Quality}

To evaluate the relationship between denoising performance and the quality of flow labels, we compare three types of supervision including synthetic ground-truth flow, estimated flow, and estimated flow with synthetic perturbations on the FANVID dataset. The ground-truth flow is derived from the known geometry and camera transformations used in 3DGS rendering. For the perturbed flow, we apply Gaussian blur and additive Gaussian noise to the estimated flow, increasing its mean endpoint error (EPE) by up to 15 pixels. As shown in Table~\ref{tab:ablation_flow}, replacing estimated flow with ground-truth flow improves the 2-step results by only 0.03 dB PSNR, 0.0007 SSIM, and 0.0036 LPIPS. Conversely, increasing the mean EPE by 15 pixels decreases performance by only 0.05 dB PSNR, 0.0020 SSIM, and 0.0052 LPIPS. These results indicate that denoising performance is not highly sensitive to flow-label accuracy. Flow supervision mainly serves as an auxiliary signal for learning sparse correspondence attention, rather than being a key determinant of restoration quality.

\begin{table}[t]
	\centering
	\caption{Effect of flow-label quality on FANVID~\cite{FANVID_ECCV24} under Poisson-Gaussian noise ($\alpha=0.03$, $\sigma=0.3$).}
	\label{tab:ablation_flow}
	\fitTableToWidth
	{
		\small
		\renewcommand{\tabcolsep}{4pt}{}
		\renewcommand{\arraystretch}{1.1}
		\begin{tabular}{cccccccc}
			\toprule
			\multirow{2}{*}{Flow label} & \multirow{2}{*}{Added mean EPE} & \multicolumn{3}{c}{1-step} & \multicolumn{3}{c}{2-step} \\
			\cmidrule(lr){3-5} \cmidrule(lr){6-8}
			 & & PSNR $\uparrow$ & SSIM $\uparrow$ & LPIPS $\downarrow$ & PSNR $\uparrow$ & SSIM $\uparrow$ & LPIPS $\downarrow$ \\
			\midrule
			Ground-truth flow & -- & 30.31 & 0.8422 & 0.2199 & 30.61 & 0.8513 & 0.2085 \\
			Estimated flow (default) & -- & 30.29 & 0.8401 & 0.2215 & 30.58 & 0.8506 & 0.2121 \\
			\midrule[0.3pt]
			\multirow{3}{*}{Perturbed estimated flow} & 5 px & 30.27 & 0.8393 & 0.2232 & 30.56 & 0.8497 & 0.2140 \\
			 & 10 px & 30.25 & 0.8382 & 0.2249 & 30.53 & 0.8487 & 0.2164 \\
			 & 15 px & 30.25 & 0.8383 & 0.2261 & 30.53 & 0.8486 & 0.2173 \\
			\bottomrule
		\end{tabular}
	}
\end{table}

\subsection{Number of Inference Steps}

Table~\ref{tab:fm_steps} compares standard FM with $1$, $2$, $4$, and $8$ solver steps and UGD-FM with one-step or two-step inference. Standard FM yields consistent gains from 1 to 4 solver steps. However, at 8 steps, PSNR and SSIM drop, while LPIPS improves, indicating that additional steps do not consistently improve restoration quality. We attribute the fidelity degradation to error accumulation during iterative forward-Euler updates, which may cause global blur and detail loss. In contrast, UGD-FM employs stage-focused and inference-consistent training for efficient inference, achieving SOTA performance with only 2 steps.

\begin{table}[t]
	\centering
	\caption{Effect of the number of inference steps on FANVID~\cite{FANVID_ECCV24} under Poisson-Gaussian noise ($\alpha=0.03$, $\sigma=0.3$).}
	\label{tab:fm_steps}
	\fitTableToWidth
	{
		\small
		\renewcommand{\tabcolsep}{4pt}{}
		\renewcommand{\arraystretch}{1.1}
		\begin{tabular}{lcccc}
			\toprule
			Method & Steps & PSNR $\uparrow$ & SSIM $\uparrow$ & LPIPS $\downarrow$ \\
			\midrule
			\multirow{4}{*}{Standard FM} & 1 & 30.28 & 0.8389 & 0.2357 \\
			 & 2 & 30.34 & 0.8401 & 0.2312 \\
			 & 4 & 30.36 & 0.8422 & 0.2291 \\
			 & 8 & 30.02 & 0.8340 & 0.2265 \\
			\midrule
			\multirow{2}{*}{UGD-FM} & 1 & 30.29 & 0.8401 & 0.2215 \\
			 & 2 & 30.58 & 0.8506 & 0.2121 \\
			\bottomrule
		\end{tabular}
	}
\end{table}

\subsection{Experiments on Misalignment Magnitudes}

We vary the misalignment magnitudes on FANVID from aligned guidance to $x=2,4,6,8$, using the same virtual-camera displacement definition as in Appendix~\ref{Details_of_datasets}. Larger $x$ indicates stronger parallax. As shown in Table~\ref{tab:misalignment_levels}, guided denoising methods like NAID~\cite{NAID_TMM24} perform well only with aligned guidance or mild misalignment. In contrast, our UGD-FM maintains strong performance as the misalignment increases, demonstrating its robustness to unaligned guidance.

\begin{table}[t]
	\centering
	\caption{Comparison under different misalignment levels on FANVID~\cite{FANVID_ECCV24} with Poisson--Gaussian noise ($\alpha=0.03$, $\sigma=0.3$).}
	\label{tab:misalignment_levels}
	\fitTableToWidth
	{
		\small
		\renewcommand{\tabcolsep}{6pt}{}
		\renewcommand{\arraystretch}{1.1}
		\begin{tabular}{clccc}
			\toprule
			Misalignment & Method & PSNR $\uparrow$ & SSIM $\uparrow$ & LPIPS $\downarrow$ \\
			\midrule
			\multirow{4}{*}{Aligned} & Restormer~\cite{Restormer_CVPR22} & 29.20 & 0.7981 & 0.3497 \\
			 & NAID~\cite{NAID_TMM24} & 31.93 & 0.8961 & 0.1616 \\
			 & UGD-FM (NAFNet, 1-step) & 31.75 & 0.8898 & 0.1771 \\
			 & UGD-FM (NAFNet, 2-step) & 32.06 & 0.8984 & 0.1535 \\
			\midrule[0.3pt]
			\addlinespace[2pt]
			\multirow{3}{*}{$x=2$}
			 & NAID~\cite{NAID_TMM24} & 29.22 & 0.8071 & 0.2919 \\
			 & UGD-FM (NAFNet, 1-step) & 30.50 & 0.8478 & 0.2129 \\
			 & UGD-FM (NAFNet, 2-step) & 30.86 & 0.8597 & 0.1988 \\
			\midrule[0.3pt]
			\addlinespace[2pt]
			\multirow{3}{*}{$x=4$}
			 & NAID~\cite{NAID_TMM24} & 29.05 & 0.7970 & 0.3324 \\
			 & UGD-FM (NAFNet, 1-step) & 30.29 & 0.8409 & 0.2250 \\
			 & UGD-FM (NAFNet, 2-step) & 30.62 & 0.8523 & 0.2119 \\
			\midrule[0.3pt]
			\addlinespace[2pt]
			\multirow{3}{*}{$x=6$}
			 & NAID~\cite{NAID_TMM24} & 28.99 & 0.7933 & 0.3540 \\
			 & UGD-FM (NAFNet, 1-step) & 30.12 & 0.8351 & 0.2351 \\
			 & UGD-FM (NAFNet, 2-step) & 30.42 & 0.8458 & 0.2237 \\
			\midrule[0.3pt]
			\addlinespace[2pt]
			\multirow{3}{*}{$x=8$}
			 & NAID~\cite{NAID_TMM24} & 28.97 & 0.7922 & 0.3673 \\
			 & UGD-FM (NAFNet, 1-step) & 29.97 & 0.8297 & 0.2448 \\
			 & UGD-FM (NAFNet, 2-step) & 30.24 & 0.8394 & 0.2352 \\
			\bottomrule
		\end{tabular}
	}
\end{table}

\subsection{More Evaluation on Real-World Datasets}

We conduct additional experiments on RNS~\cite{RNS_CVPR25}, which is a real-world dataset captured using an RGB-NIR dual-camera system. We use the original images without epipolar rectification, with RGB images from the right camera as targets and NIR images from the left camera as guidance. Table~\ref{tab:cmp_RNS} reports results under Poisson-Gaussian noise ($\alpha=0.03$, $\sigma=0.3$). UGD-FM achieves the best performance, demonstrating its generalizability to real-world data.

\begin{table}[t]
	\centering
	\caption{Quantitative comparison on the RNS~\cite{RNS_CVPR25} dataset under Poisson-Gaussian noise ($\alpha=0.03, \sigma=0.3$). The \textbf{best} and \underline{second-best} results are highlighted in bold and underlined, respectively.}
	\label{tab:cmp_RNS}
		\fitTableToWidth
		{
			\small
			\renewcommand{\tabcolsep}{4pt}{}
			\renewcommand{\arraystretch}{1.1}
			\begin{tabular}{lccc}
				\toprule
				Method & PSNR $\uparrow$ & SSIM $\uparrow$ & LPIPS $\downarrow$ \\
				\midrule
				Restormer~\cite{Restormer_CVPR22} & 28.20 & 0.7761 & 0.3793 \\
				RDBM~\cite{RDBM_CVPR26} & 28.38 & 0.7678 & 0.3377 \\
                \midrule
				NAID~\cite{NAID_TMM24} & 28.56 & 0.7997 & 0.3137 \\
                SGDFormer~\cite{SGDFormer_IF25} & 28.39 & 0.7863 & 0.3307 \\
				RFfusion~\cite{RFfusion_NIPS2025} & 27.76 & 0.7626 & 0.3144 \\
				3M-TI~\cite{3MTI_CVPR2026} & 28.62 & 0.8011 & 0.3024 \\
				\midrule
				UGD-FM (NAFNet, 1-step) & \underline{28.96} & \underline{0.8110} & \underline{0.2505} \\
				UGD-FM (NAFNet, 2-step) & \textbf{29.10} & \textbf{0.8177} & \textbf{0.2280} \\
				\bottomrule
			\end{tabular}
		}
\end{table}

\subsection{Inference Efficiency}
Table~\ref{tab:inference} reports inference time and peak memory usage at an input resolution of $768\times1024$ on a single NVIDIA RTX 4090 GPU. UGD-FM achieves state-of-the-art performance without incurring prohibitive memory or inference-time overhead.

\begin{table}[t]
	\centering
	\caption{Inference efficiency at input resolution $768\times1024$ on an NVIDIA RTX 4090 GPU.}
	\label{tab:inference}
	\fitTableToWidth
	{
		\small
		\renewcommand{\tabcolsep}{6pt}{}
		\renewcommand{\arraystretch}{1.1}
		\begin{tabular}{lcc}
			\toprule
			Method & Time (ms) & Peak memory (GiB) \\
			\midrule
			MPRNet~\cite{MPRNet_CVPR21} & 437 & 6.3 \\
			NAFNet~\cite{NAFNet_ECCV22} & 76 & 0.9 \\
			Uformer~\cite{Uformer_CVPR22} & 171 & 4.7 \\
			Restormer~\cite{Restormer_CVPR22} & 676 & 6.7 \\
			Xformer~\cite{Xformer_ICLR24} & 1390 & 7.8 \\
			RDBM~\cite{RDBM_CVPR26} & 3431 & 9.3 \\
			\midrule
			CUNet~\cite{CUNet_TPAMI20} & 162 & 1.9 \\
			DVN~\cite{DVN_AAAI22} & 159 & 3.9 \\
			RFFNet~\cite{RFFNet_MM24} & 164 & 11.4 \\
			NAID~\cite{NAID_TMM24} & 120 & 1.5 \\
			SANet~\cite{SANet_CVPR23} & 194 & 4.4 \\
			RFfusion~\cite{RFfusion_NIPS2025} & 841 & 20.7 \\
			3M-TI~\cite{3MTI_CVPR2026} & 654 & 9.7 \\
			\midrule
			UGD-FM (NAFNet, 1-step) & 250 & 5.1 \\
			UGD-FM (NAFNet, 2-step) & 487 & 5.1 \\
			UGD-FM (Uformer, 1-step) & 441 & 5.0 \\
			UGD-FM (Uformer, 2-step) & 829 & 5.0 \\
			UGD-FM (Restormer, 1-step) & 368 & 5.0 \\
			UGD-FM (Restormer, 2-step) & 694 & 5.0 \\
			\bottomrule
		\end{tabular}
	}
\end{table}

\subsection{Visualization of the Mutually Reinforcing Process}

\begin{figure}[t]
    \centering
    \includegraphics[width=\linewidth]{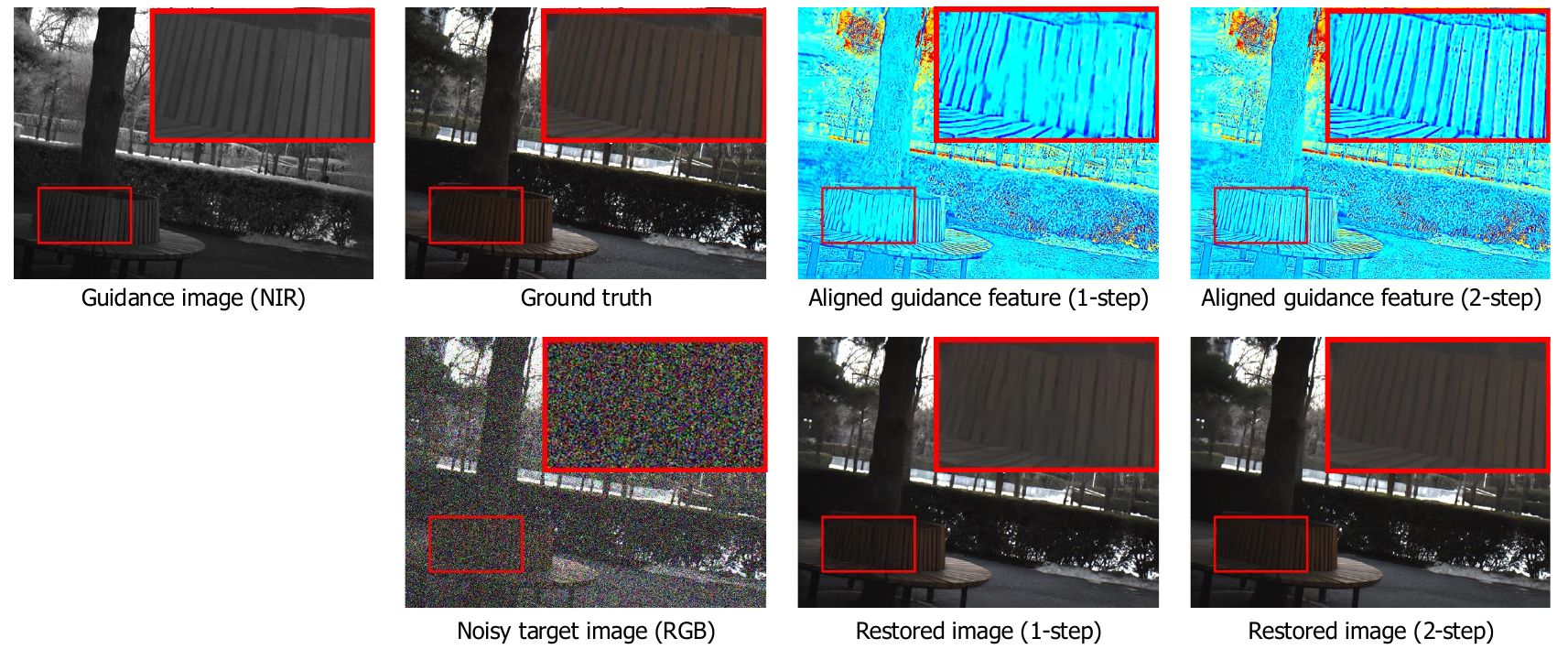}
    \includegraphics[width=\linewidth]{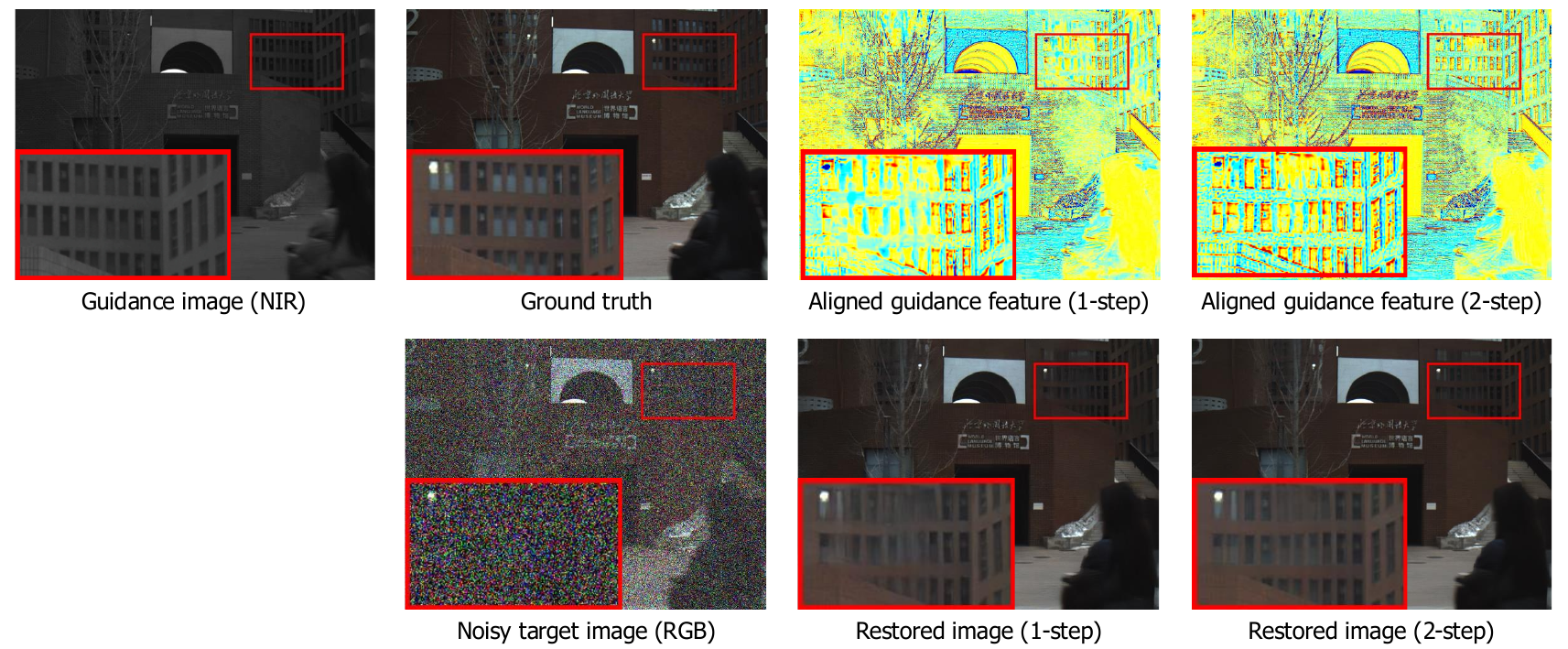}
    \caption{Visualization of the mutually reinforcing process between image restoration and guidance aggregation.}
    \label{fig:vis_feature}
\end{figure}

Figure~\ref{fig:vis_feature} demonstrates the mutually reinforcing process between image restoration and guidance aggregation. The visualized features correspond to the guidance features of the finest level of decoding blocks. As the flow matching trajectory evolves, the guidance features become more structured and coherent.

\subsection{More Qualitative Comparisons}

\begin{figure}[t]
    \centering
    \includegraphics[width=\linewidth]
    {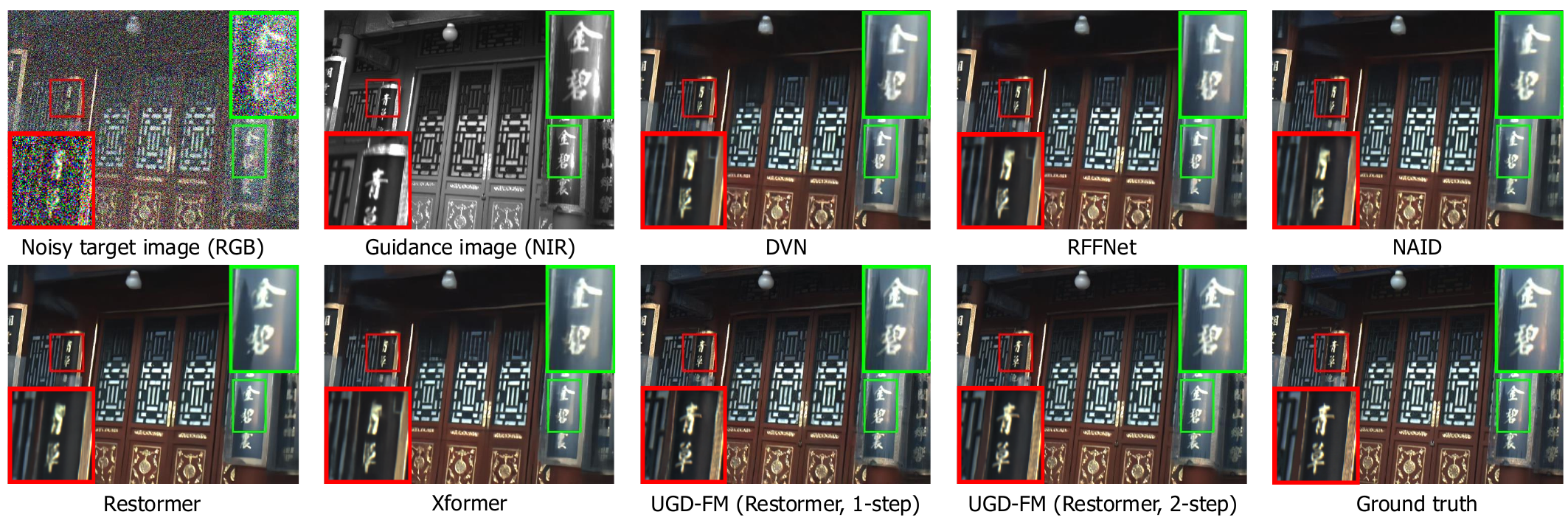}
    \includegraphics[width=\linewidth]
    {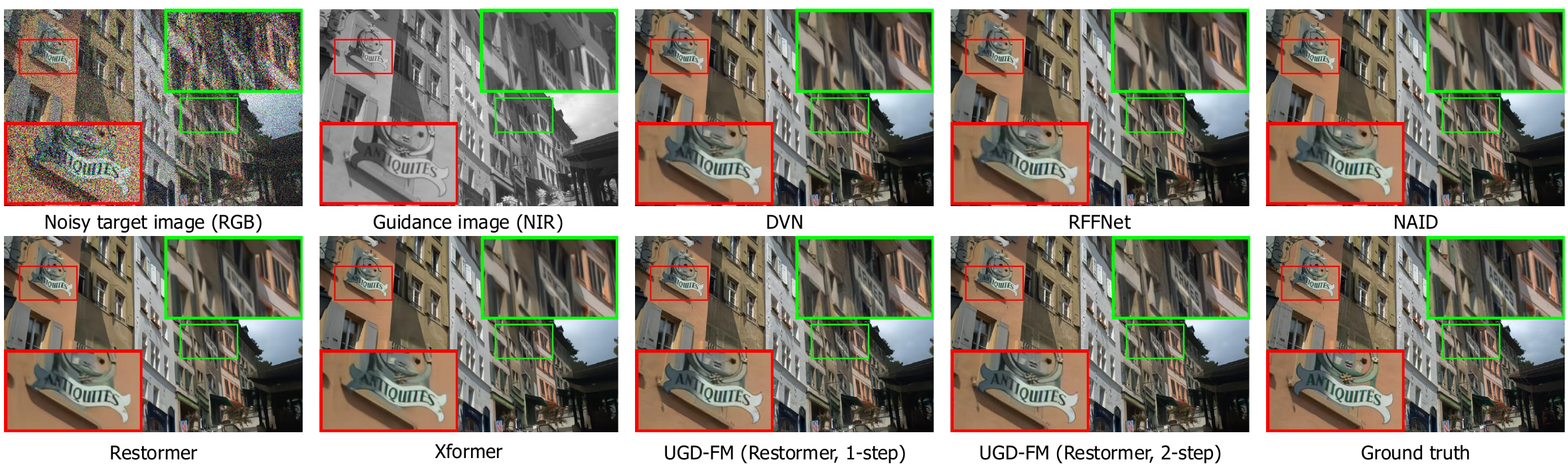}
    \includegraphics[width=\linewidth]
    {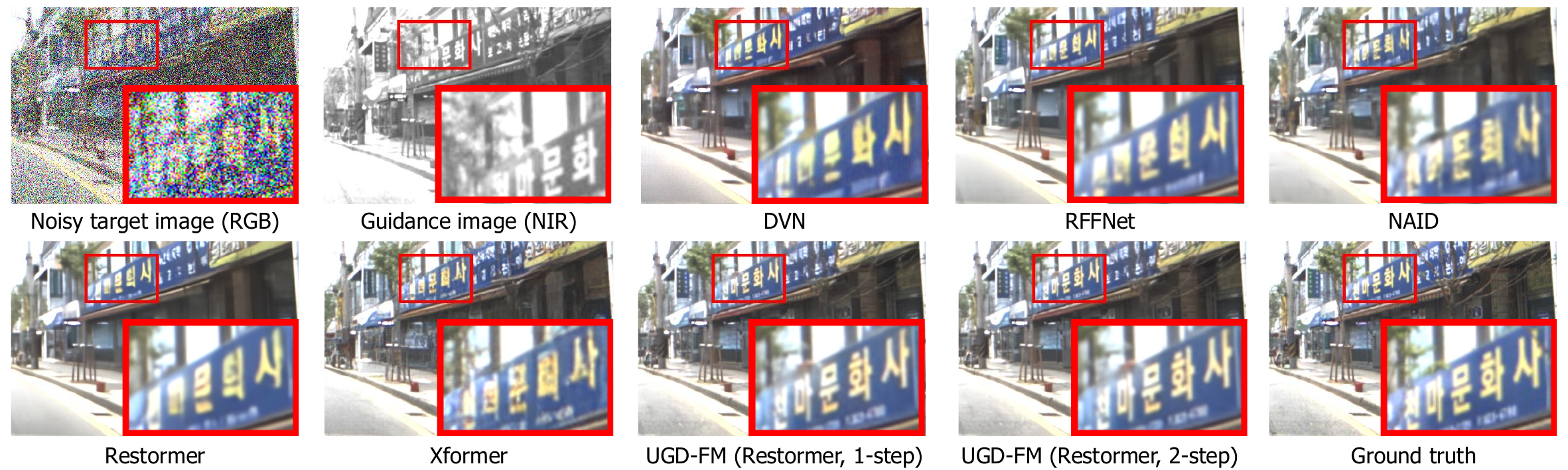}
    \caption{More qualitative comparison of image denoising results across different approaches.}
    \label{fig:vis_cmp_supp}
\end{figure}

We provide more qualitative comparisons of different approaches in Fig.~\ref{fig:vis_cmp_supp}. Single image denoising approaches generally cannot preserve structures and textures well. Among all guided denoising methods, our UGD-FM produces denoised images with more salient structures and details.

\section{Limitations}

\begin{figure}[t]
    \centering
    \includegraphics[width=\linewidth]{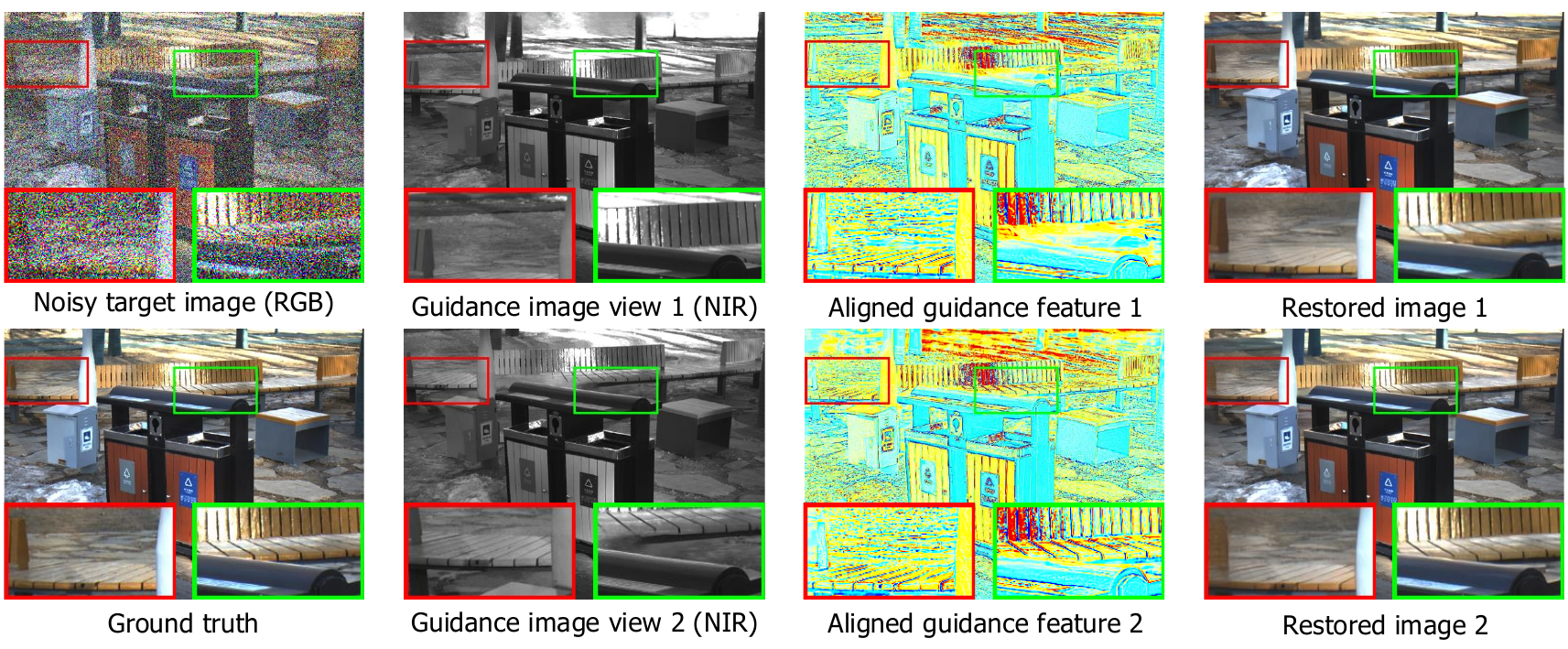}
    \caption{Impact of parallax-induced occlusion on guided denoising. The top and bottom rows present visualizations of the same sample under different parallax conditions. Red and green boxes highlight how occlusion affects feature alignment and reconstruction quality.}
    \label{fig:limit}
\end{figure}

UGD-FM remains challenged by severe 3D parallax, where occlusions and incomplete cross-view observations can make local guidance ambiguous or unreliable, as shown in Fig.~\ref{fig:limit}. Joint restoration and cross-modal alignment also introduce computational overhead. Despite these limitations, we believe UGD-FM provides a practical solution for unaligned guided image denoising.

\section{Broader Impacts}

UGD-FM can benefit low-light photography through image restoration using cross-modal guidance. However, its use in unauthorized surveillance may raise privacy concerns, and restored images may contain plausible but inaccurate details. Deployment in safety-critical applications therefore requires careful validation and further assessment of uncertainty and reliability.

\end{document}